\documentclass[10pt,twocolumn,letterpaper]{article}

\usepackage{cvpr}
\usepackage{times}
\usepackage{multirow}
\usepackage{epsfig}
\usepackage{graphicx}
\usepackage{amsmath}
\usepackage{amssymb}
\usepackage[numbers,sort]{natbib}
\usepackage{booktabs}
\usepackage[font=small,labelfont=bf]{caption}
\usepackage[pagebackref=true,breaklinks=true,letterpaper=true,colorlinks,bookmarks=false]{hyperref}
\usepackage{lib}
\cvprfinalcopy % *** Uncomment this line for the final submission

\newcommand{\addpic}[1]{\includegraphics[width=8em]{#1}}
\newcommand{\addpichalf}[1]{\includegraphics[width=4em]{#1}}
\input{shortbold.sty}
\def\cvprPaperID{1622} % *** Enter the CVPR Paper ID here

\begin{document}

%%%%%%%%% TITLE
% \title{Computer Vision goes to the Detroit Zoo!}
%\title{Learning Canonical Surface Mapping and Articulation \\ via Geometric Cycle Consistency}
%\title{Making Androids Dream of Articulated Sheep}
\title{Articulation-aware Canonical Surface Mapping}
% title: 
% Articulation-aware Canonical Surface Mapping
% Joint Learning of __ and __
% Learning Canonical Surface Mapping for Articulated Object Categories
% {Learning} Canonical Surface Mapping and Articulation via Geometric Cycle Consistency
% Canonical Surface Mapping via Articulation-aware Geometric Cycle Consistency
% Making Androids Dream of Articulated Sheep

\author{
Nilesh Kulkarni$^1$ \qquad Abhinav Gupta$^{2,3}$ \qquad David F. Fouhey$^1$ \qquad Shubham Tulsiani$^3$\\
$\,^1$University of Michigan \qquad$\,^2$Carnegie Mellon University \qquad$\,^3$Facebook AI Research\\
{\tt \small \{nileshk, fouhey\}@umich.edu} \qquad   \tt \small abhinavg@cs.cmu.edu \tt \small shubhtuls@fb.com
\\ 
{\tt \small \href{https://nileshkulkarni.github.io/acsm/}{https://nileshkulkarni.github.io/acsm/}}
}
% \maketitle
%\thispagestyle{empty}
% \twocolumn[{%
% \renewcommand\twocolumn[1][]{#1}%
% \vspace{-4em}
% \maketitle
% \vspace{-3em}
% \begin{center}
%   \centering \includegraphics[width=\textwidth]{figures/teaser.pdf} \captionof{figure}{We tackle the tasks of: a) canonical surface mapping (CSM) \ie mapping pixels to corresponding points on a template shape, and b) predicting articulation of this template. Our approach allows learning these without relying on keypoint supervision, and we visualize the results obtained across several categories. The color across the template 3D model on the left and image pixels represent the predicted mapping among them, while the smaller 3D meshes represent our predicted articulations in camera (top) or a novel (bottom) view.}
%   \{teaser}
% \end{center}%figlabel
% }]

\twocolumn[{%
\renewcommand\twocolumn[1][]{#1}%
\vspace{-4em}
\maketitle
\vspace{-3em}
\begin{center}
   \centering \includegraphics[width=\textwidth]{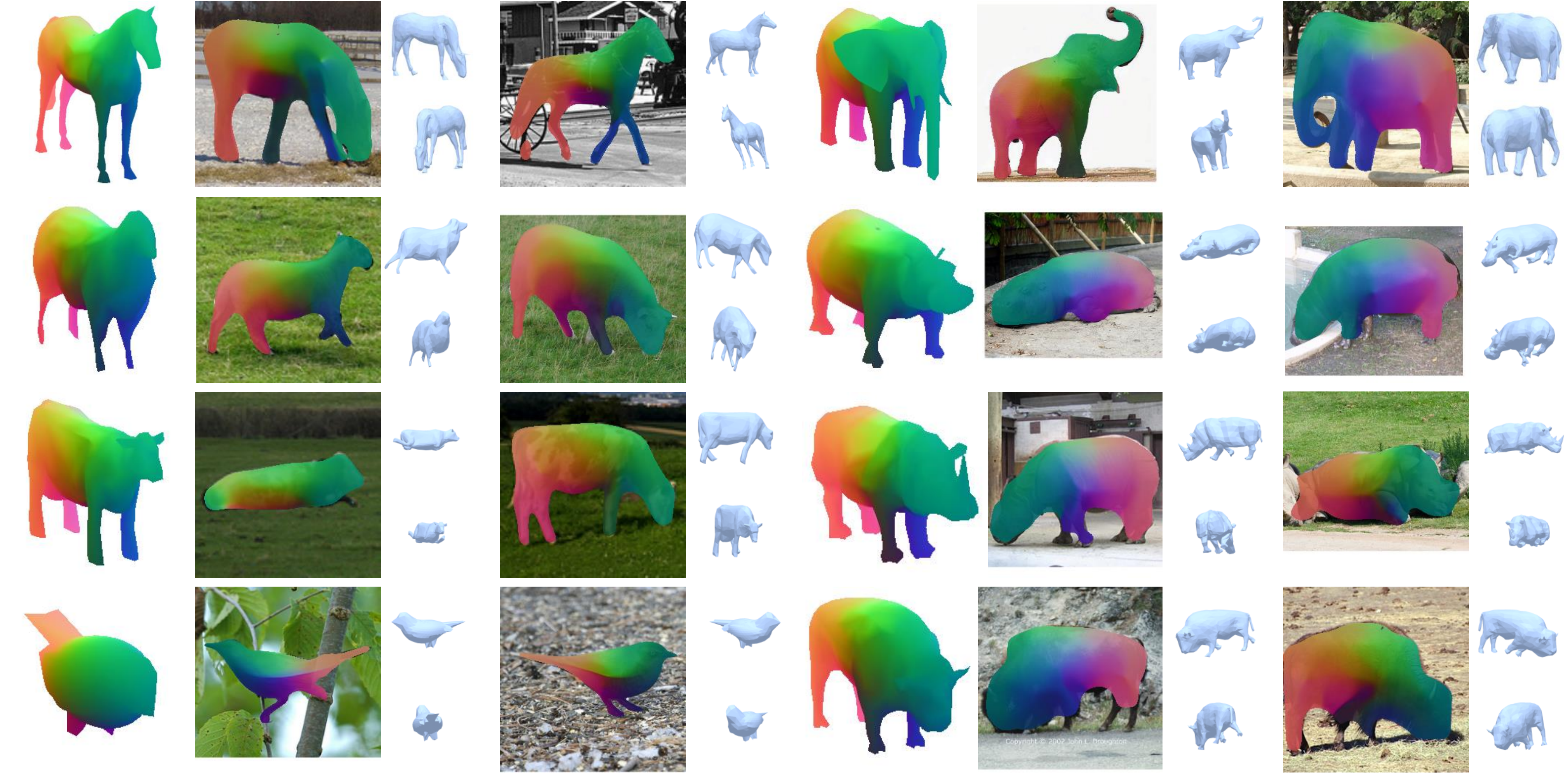} \captionof{figure}{We tackle the tasks of: a) canonical surface mapping (CSM) \ie mapping pixels to corresponding points on a template shape, and b) predicting articulation of this template. Our approach allows learning these without relying on keypoint supervision, and we visualize the results obtained across several categories. The color across the template 3D model on the left and image pixels represent the predicted mapping among them, while the smaller 3D meshes represent our predicted articulations in camera (top) or a novel (bottom) view.}
   \figlabel{teaser}
\end{center}%
}]
%%%%%%%%% ABSTRACT
\begin{abstract}
\vspace{-2mm}
We tackle the tasks of: 1) predicting a Canonical Surface Mapping (CSM) that indicates the mapping from 2D pixels to corresponding points on a canonical template shape , and 2) inferring the articulation and pose of the template corresponding to the input image. While previous approaches rely on keypoint supervision for learning, we present an approach that can learn without such annotations. Our key insight is that these tasks are geometrically related, and we can obtain supervisory signal via enforcing consistency among the predictions. We present results across a diverse set of animal object categories, showing that our method can learn articulation and CSM prediction from image collections using only foreground mask labels for training. We empirically show that allowing articulation helps learn more accurate CSM prediction, and that enforcing the consistency with predicted CSM is similarly critical for learning meaningful articulation.
\end{abstract}

%%%%%%%%% BODY TEXT
\section{Introduction}

We humans have a remarkable ability to associate our 2D percepts with 3D concepts, at both a global and a local level. As an illustration, given a pixel around the nose of the horse depicted in \figref{teaser} and an abstract 3D model, we can easily map this pixel to its corresponding 3D point.  Further, we can also understand the global relation between the two, \eg the 3D structure in the image corresponds to the template with the head bent down. In this work, we pursue these goals of a local and global 3D understanding, and tackle the tasks of: a) \emph{canonical surface mapping} (CSM) \ie mapping from 2D pixels to a 3D template, and b) predicting this template's \emph{articulation} corresponding to the image.

While several prior works do address these tasks, they do so independently, typically relying on large-scale annotation for providing supervisory signal. For example, Guler \etal~\cite{alp2018densepose} show impressive mappings from pixels to a template human mesh, but at the cost of hundreds of thousands of annotations. Similarly, approaches pursuing articulation inference~\citep{cmrKanazawa18, zuffi2019three} also rely on keypoint annotation to enable learning. While these approaches may be used for learning about categories of special interest \eg humans, cats \etc, the reliance on such large-scale annotation makes them unscalable for generic classes. In contrast, our goal in this work is to enable learning  articulation and pixel to surface mappings without leveraging such manual annotation. 

Our key insight is that these two forms of prediction are in fact geometrically related. The CSM task yields a dense local mapping from pixels to the template shape, and conversely, inferring the global articulation (and camera pose) indicates a transform of this template shape onto the image. We show that these two predictions can therefore provide supervisory signal for each other, and that enforcing a consistency between them can enable learning \emph{without requiring direct supervision} for either of these tasks. We present an approach that operationalizes this insight, and allows us to learn CSM and articulation prediction for generic animal object categories from online image collections.

We build on upon our prior work~\cite{kulkarni2019canonical} that, with a similar motivation, demonstrated that it is possible to learn CSM prediction without annotation, by relying on the consistency between rigid reprojections of the template shape and the predicted CSM.  However, this assumed that the object in an image is rigid \eg does not have a bent head, moving leg \etc, and this restricts the applicability and accuracy for objects that exhibit articulation. In contrast, we explicitly allow predicting articulations, and incorporate these before enforcing such consistency, and our approach thereby: a) helps us learn articulation prediction without supervision, and b) leads to more accurate CSM inference. We present qualitative and quantitative results across diverse classes indicating that we learn accurate articulation and pixel to surface mappings across these. Our approach allows us to learn using ImageNet ~\cite{deng2009imagenet} images with approximate segmentations from off-the-shelf systems, thereby enabling learning in setups that previous supervised approaches could not tackle, and we believe this is a step towards large-internet-scale 3D understanding.
\section{Related Work}

\noindent \textbf{Pose and Articulation Prediction.} One of the tasks we address is that of inferring the camera pose and articulation corresponding to an input image. The task of estimating pose for rigid objects has been central to understanding objects in 3D scenes, and addressed by several works over the decades, from matching based methods~\cite{huttenlocher1990recognizing,pepik2012teaching,xiang2014beyond}, to recent CNN based predictors~\cite{tulsiani2015viewpoints,su2015render}. Closer to our work, a natural generalization of this task towards animate objects is to also reason about their articulation \ie movement of parts, and a plethora of fitting based~\cite{kanazawa2016learning,bogo2016keep} or prediction based~\cite{zuffi2019three,xiang2019monocular,hmrKanazawa17} methods have been proposed to tackle this. While these show impressive results across challenging classes, these methods crucially rely on (often dense) 2D keypoint annotations for learning, and sometimes even inference. Our goal is to learn such a prediction without requiring this supervision. We show that enforcing consistency with a dense pixel to 3D mapping allows us to do so.

\vspace{1mm}
\noindent \textbf{Dense Mappings and Correspondences.} 
In addition to learning articulation, we a predict per-pixel mapping to a template shape. Several previous approaches similarly pursue pixel to surface~\cite{alp2018densepose, zhu2016face,neverova2019slim, sinha2017surfnet, maron2017convolutional} or volume~\cite{Wang_2019_CVPR} mappings, but unlike our approach, crucially rely on direct supervision towards this end. Note that these mappings also allow one to recover correspondences across images, as corresponding pixels have similar representations. Towards this general goal of learning representations that respect correspondence, several prior works attempt to design~\cite{lowe2004sift}, or learn features invariant to camera movement~\cite{xiang2017posecnn,florence2018dense}, or synthetic transforms~\cite{thewlis2017unsupervised}. While the latter approaches can be leveraged without supervision, the embedding does not enforce a geometric structure, which is what crucially helps us jointly learn articulation and pose. Closer to our work, Kulkarni \etal~\cite{kulkarni2019canonical} learn a similar mapping without direct supervision but unlike us, ignore the effects of articulation, which we model to obtain more accurate results.

\vspace{1mm}
\noindent \textbf{Reconstructing Objects in 3D.} Our approach can be considered as predicting a restricted form of 3D reconstruction from images, by `reconstructing' the 3D shape in the form of an articulated template shape and its pose. There are many existing approaches which tackle more general forms of 3D prediction, ranging from volumetric prediction~\cite{choy2016universal,girdhar2016learning} to point cloud inference~\cite{fan2017point,lin2018learning}. Perhaps more directly related to our representation is the line of work that, following the seminal work on Blanz and Vetter~\cite{BlanzVetter}, represents the 3D in the form of a morphabble model, jointly capturing articulation and deformation~\cite{oberweger2015hands, SMPL, pavlakos2019expressive}. While all these approaches yield more expressive 3D than our approach, they typically rely on 3D supervision for training. Even methods that attempt to relax this~\cite{yan2016perspective,cmrKanazawa18,drcTulsiani17} need to leverage multi-view or keypoint supervision for learning, and in this work, we attempt to also relax this requirement.

\section{Approach}
\noindent
Given an input image $I$, our goal is to infer: (1) a per-pixel correspondence $C$, mapping each pixel in $I$ to a point on the template; (2) an articulation $\delta$ of the 3D template, as well as a camera pose $\pi = (s,\RB,\tB)$ that represents how the object appears in or projects into the image.
We operationalize this with two deep networks $f_\theta$ and $g_{\theta'}$ that take as input image $I$, and produce $C \equiv f_\theta(I)$ and $\delta,\pi \equiv g_{\theta'}(I)$ respectively. Instead of requiring large-scale manual keypoint annotations for learning these mappings, we strive for an approach that can learn without such keypoint labels, using only category-level image collections with (possibly noisy) foreground masks. 
Our key insight is that the two tasks of predicting pixel to 3D template mappings and transformations of template to image frame are geometrically related, and we can enforce consistency among the predictions to obtain supervisory signal for both. Recent work by Kulkarni \etal~\cite{kulkarni2019canonical} leveraged a similar insight to learn CSM prediction, but assumed a rigid template, which is a fundamentally limiting assumption for most animate object classes. We present an approach that further allows the model to articulate, and observe that this enables us to both learn about articulation without supervision, and recover more accurate pixel to surface mappings.

The core loss and technique is a geometric consistency loss that synchronizes the CSM, articulation and pose,  which we present along with our articulation parametrization in \secref{gcc}. We then describe how we train $f_\theta$ and $g_{\theta'}$ in \secref{learning}, which builds on this core loss by adding auxiliary losses based on mask supervision and shows how our approach can be extended to incorporate sparse keypoint supervision if available. 

\vspace{2mm}
\noindent {\bf Mesh Preliminaries.} We note that the surface of a mesh is a 2D manifold in 3D space and we can therefore construct a 2D parametrization of a 3D surface as $\phi: [0,1)^2 \rightarrow S$. This maps a 2D vector $\mathbf{u}$ to a unique point on the surface of the template shape $S$. Given such a surface parametrization, a canonical surface mapping $C$ is defined as a 2D vector image, such that for a given pixel $\mathbf{p}$, $\phi(C[\mathbf{p}])$ is its corresponding 3D point on the template. Please see the supplemental for additional details on constructing $\phi$ for a template shape.

\subsection{Articulation-aware Geometric Consistency}
\seclabel{gcc}
\vspace{2mm}
\noindent {\bf Articulation Parametrization.}
Given a template shape in the form of a mesh, we approximately group its vertices into functional `parts' \eg head, neck, legs \etc, as well as define a hierarchy among these parts. While our initial grouping is discrete, following standard practices in computer graphics~\cite{lewis2000pose}, we `soften' the per-vertex assignment as depicted in \figref{smooth_alpha}. Assuming $K$ parts, this `rigging' procedure yields, for each mesh vertex $v$, the associated memberships $\alpha^v_k \in [0,1]$ corresponding to each part. Note that this annotation procedure is easily scalable, requiring only a few minutes per category (for a non-expert annotator).
\begin{figure}[t]
    \centering
    \includegraphics[width=0.40\textwidth ]{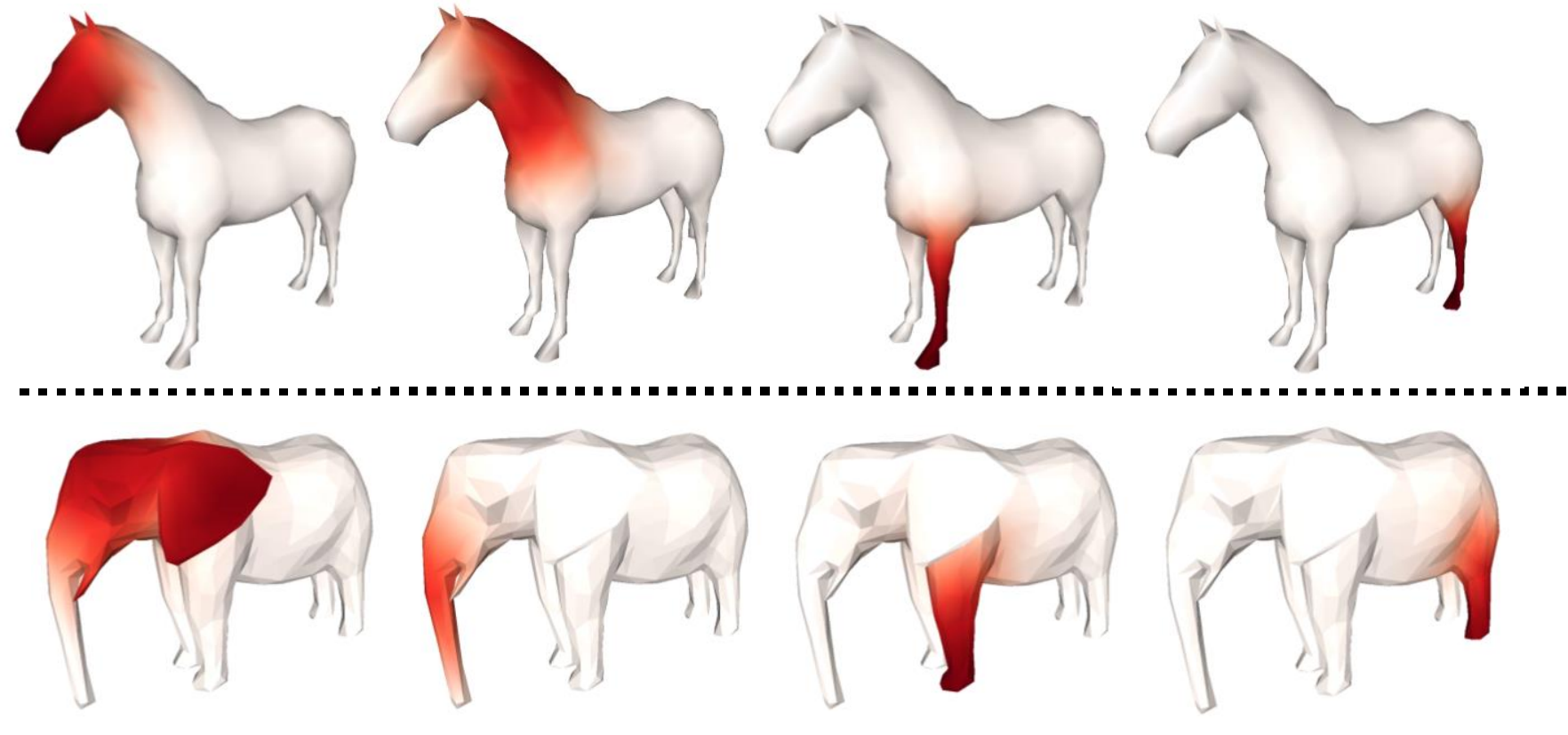}
    \caption{\textbf{Sample per-part vertex assignments.} We show softened per-vertex assignment to various parts of  quadrupeds. This pre-computed soft assignment enables us to obtain smooth deformations of the template mesh  across the part boundaries under articulation.
    }
    \figlabel{smooth_alpha}
\end{figure}
The articulation $\delta$ of this template is specified by a rigid transformation (translation and rotation) of each part w.r.t. its parent part \ie $\delta \equiv \{(t_k, R_k)\}$, with the `body' being the root part. Given (predicted) articulation parameters $\delta$, we can compute a global transformation $\mathcal{T}_k(\cdot, \delta)$ for each part, s.t. a point $p$ on the part in the canonical template moves to $\mathcal{T}_k(p, \delta)$ in the articulated template (see supplemental for details). Therefore, given a vertex $v$ on the canonical template mesh, we can compute its position after articulation as $\sum_k \alpha^v_k~ \mathcal{T}_k(v, \delta)$. We can extend this definition for any point $p$ on the surface using barycentric interpolation (see supplemental). We slightly overload notation for convenience, and denote by $\delta(p)$ the position of any point $p \in S$ after undergoing articulation specified by $\delta$.

\begin{figure}[t]
    \centering
    \includegraphics[width=0.45\textwidth ]{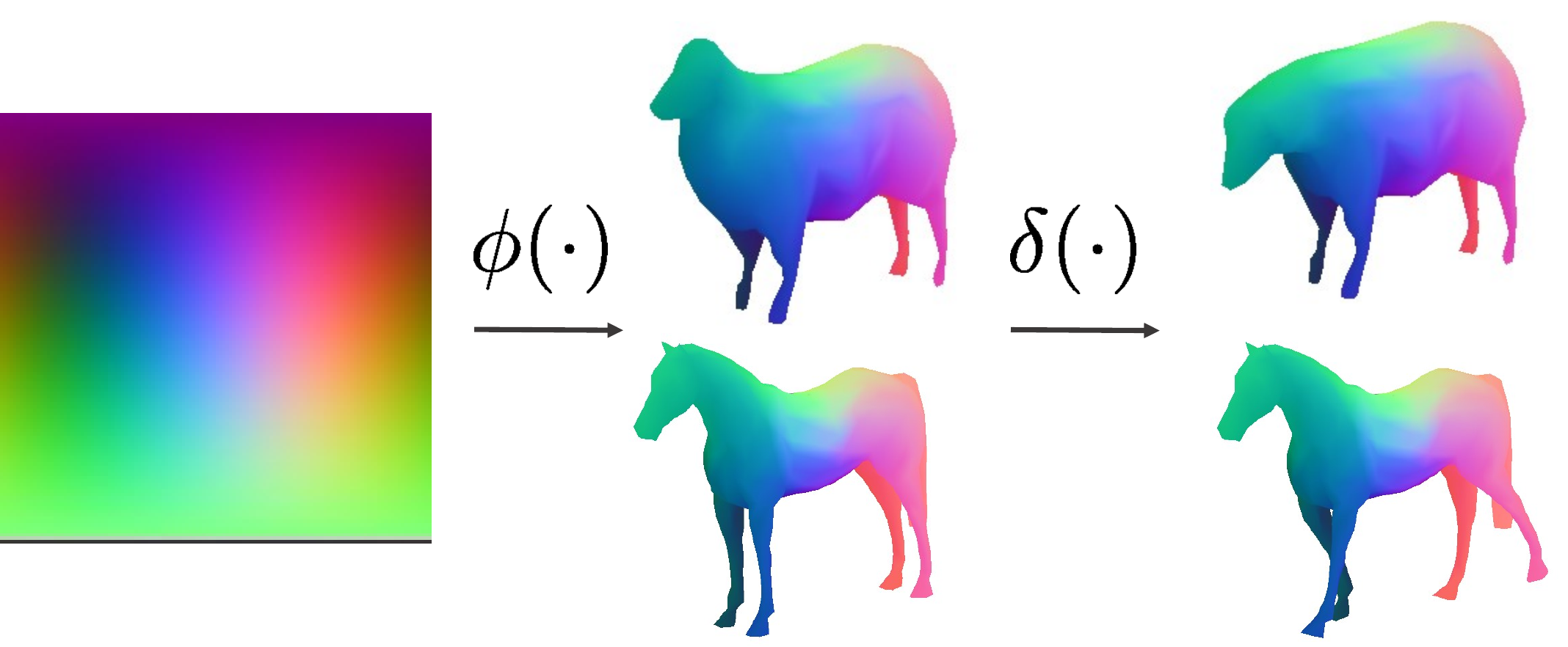}
    \caption{\textbf{Illustration of surface parametrization and articulation.} Given a 2D coordinate $\mathbf{u} \in [0,1]^2$, the function $\phi$ maps it to the surface of a template shape, which can then be  transformed according to the articulation  $\delta$ specified. We depict here the mappings from this 2D space to the articulated shapes for two possible articulations: horse with moving legs, and sheep with a bent head.} 
    \figlabel{uv_map}
\end{figure}

\vspace{2mm}
\noindent {\bf Canonical to Articulated Surface Mapping.}
For any 2D vector $\mathbf{u} \in [0,1)^2$, we can map it to the template shape via $\phi$. If the shape has undergone an articulation specified by $\delta$, we can map this vector to a point on the \emph{articulated} shape by composing the articulation and mapping, or $\delta(\phi(\mathbf{u}))$. We depict this in \figref{uv_map}, and show the mapping from the 2D space to the template under various articulations. Given a pixel to canonical surface mapping $C$, we can therefore recover for a pixel $\mathbf{p}$ its corresponding point on the articulated shape as $\delta(\phi(C[\mathbf{p}]))$.

\begin{figure}[t]
    \centering
    \includegraphics[width=0.40 \textwidth ]{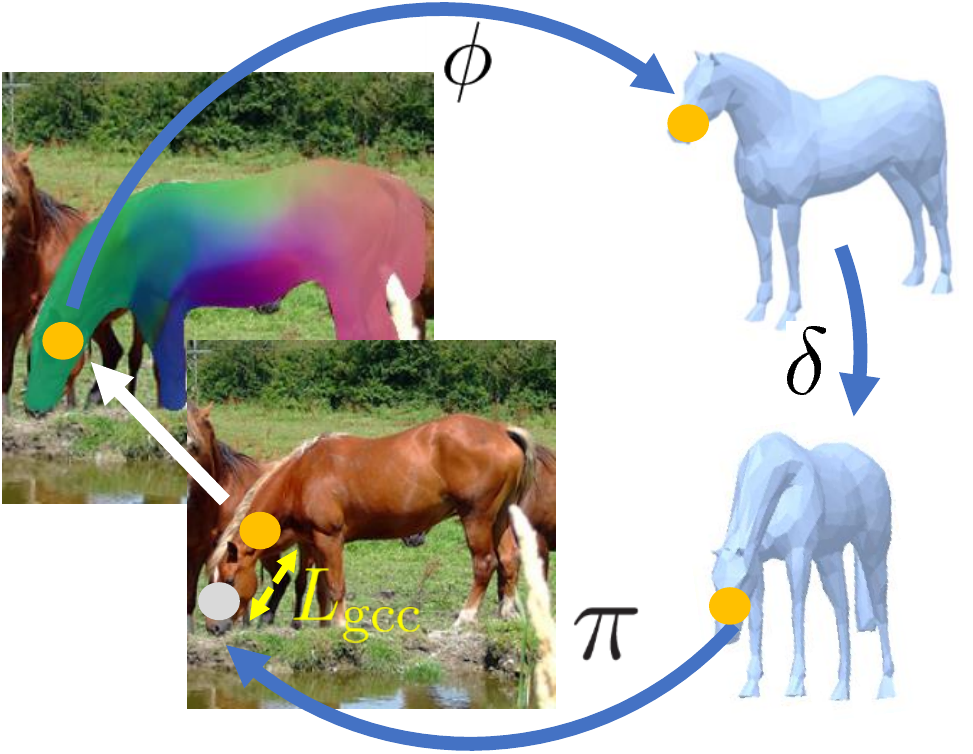}
    \caption{\textbf{Articulation-aware Geometric Cycle Consistency.} Given an image pixel, we can map it to a point on the surface of the template shape using the predicted CSM mapping and $\phi$. We then articulate the surface using $\delta$ to map points on the surface of template shape to the articulated shape. The inconsistency arising from the reprojection of points from articulated shape under the camera $\pi$ yields the geometric cycle consistency loss, $L_{\text{gcc}}$.}
    \figlabel{gcc}
\end{figure}

\vspace{2mm}
\noindent {\bf Geometric Consistency.}
The canonical surface mapping defines a 2D $\rightarrow$ 3D mapping from a pixel to a  point on the 3D mesh; we show how to use cameras observing the mesh to define a cycle-consistent loss from each mesh point to a pixel. In particular, the canonical surface mapping $C$ maps pixels to the corresponding 3D points on the (un-articulated) template. In the other direction, a (predicted) articulation $\delta$ and camera parameters $\pi$ define a mapping from this canonical shape to the image space: the mesh deforms and is then projected back into the image. 
Ideally, for any pixel $\mathbf{p}$, this 3D mapping to the template followed by articulation and projection should yield the original pixel location if the predictions are geometrically consistent. We call this constraint as geometric cycle consistency (GCC).

\vspace{2mm}
We can operationalize this to measure the inconsistency between a canonical surface mapping $C$, articulation $\delta$ and camera $\pi$, as shown in \figref{gcc}. Given a foreground pixel $\mathbf{p}$, its corresponding point on the template shape can be computed as $\phi(C[\mathbf{p}])$, and on the articulated shape as $\delta(\phi(C[\mathbf{p}]))$. Given the (predicted) camera $\pi$, we can compute its reprojection in the image frame as $\pi(\delta(\phi(C[\mathbf{p}])))$. We then penalize the difference between the initial and the reprojected pixel location to enforce consistency.
\begin{align}
    L_{\text{gcc}} = \sum_{{\bf p} \in I_{f}} \norm{{\bf p} - {\bf \bar{p}}} \quad  ; \quad {\bf \bar{p}} = \pi(\delta(\phi(C[{\bf p}])))
    \eqlabel{gcc}
\end{align}

\subsection{Learning CSM and Articulation Prediction}
\seclabel{learning}
\noindent
Recall that our goal is to train a predictor $f_\theta$ that predicts the CSM $C$ and a predictor $g_{\theta'}$ that predicts the articulation $\delta$ and camera $\pi$. Our approach, as illustrated in \figref{approach}, learns these using $L_\textrm{gcc}$ that enforces consistency among the predictions. We additionally have to add auxiliary losses based on foreground mask supervision to prevent trivial or degenerate solutions. These losses penalize the discrepancy between the annotated masks and masks rendered from the articulated mesh. We describe the learning procedure and objectives in more detail below, and then discuss incorporating keypoint supervision if available.

\begin{figure}[t]
    \centering
    \includegraphics[width=0.45\textwidth ]{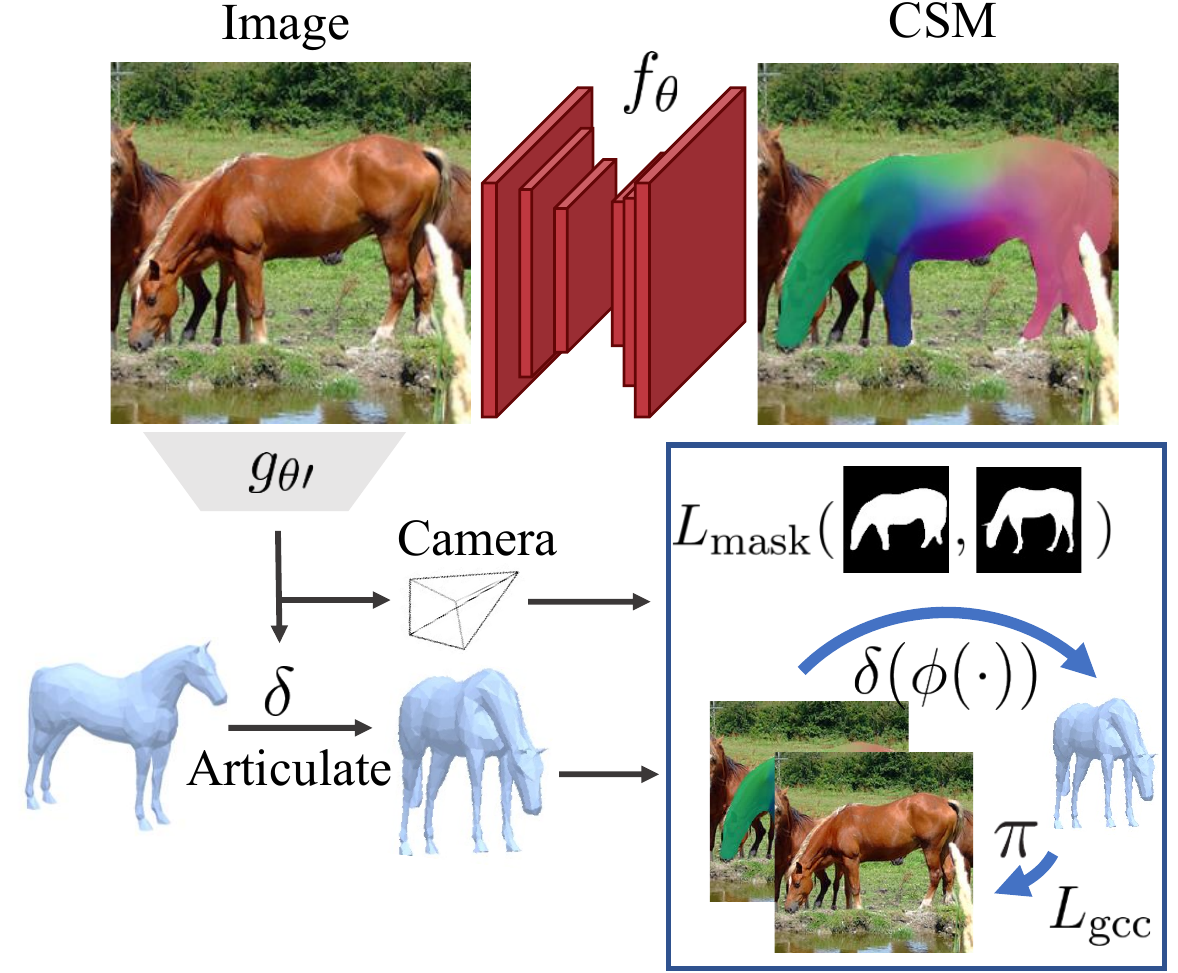}
   \caption{\textbf{Overview of our approach}. Our approach A-CSM jointly learns to predict the CSM mapping, a camera, and the articulation. We require that these predictions be consistent with each other by enforcing the $L_{\text{cyc}}$ and $L_{\text{mask}}$ constraint.}
   \figlabel{approach}
\end{figure}

\vspace{2mm}
\noindent \textbf{Visibility Constraints.} 
The GCC reprojection can be consistent even under mappings to an occluded region \eg if the pixel considered in \figref{gcc} were mapped to the other side of the horse's head, its image reprojection map still be consistent. To discourage such mappings to invisible regions, we follow Kulkarni \etal~\cite{kulkarni2019canonical} and incorporate a visibility loss $L_{\text{vis}}$ that penalizes inconsistency between the reprojected and rendered depth (for more details see supplemental).

\vspace{2mm}
\noindent \textbf{Overcoming Ambiguities via Mask Supervision.} Simply enforcing self-consistency among all predictions in absence of any grounding however, can lead to degenerate solutions. Hence, we leverage the foreground mask obtained under camera ($\pi$) for the template shape after articulation ($\delta$) to match the annotated foreground mask. As we want to encourage more precise articulation, we find it beneficial to measure the difference between the 2D distance fields induced by the foreground masks instead of simply comparing the per-pixel binary values, and define an objective $L_{\text{mask}}$ to capture this. This objective is sum of mask-consistency and mask-coverage objectives as defined in ~\cite{kar2015category}. We describe it further detail in the supplemental.

\vspace{2mm}
\noindent \textbf{Learning Objective.} Our overall training objective $L_{\text{total}}$ then minimizes a combination of the above losses:
\begin{align}
    L_{\text{total}} = L_{\text{gcc}} + L_{\text{vis}} + L_{\text{mask}}
    \eqlabel{training}
\end{align}
Additionally, instead of learning a camera and deformation predictor $g_{\theta'}$ that predicts a unique output, we follow previous approaches~\cite{mvcTulsiani18, kulkarni2019canonical, insafutdinov2018unsupervised} to learn a multi-hypothesis predictor, that helps overcome local minima. Concretely, $g_{\theta'}(I)$ outputs $8$ (pose, deformation) hypotheses, $\{(\pi_i,\delta_i)\}$, and an associated probability $c_i$, and we minimize the expected loss across these.

\vspace{2mm}
\noindent \textbf{Leveraging Optional Keypoint (KP) Supervision.} While we are primarily interested in learning without any manual keypoint annotations, our approach can be easily extended to additional annotations for some set of semantic keypoints \eg nose, left eye \etc are available. To leverage these, we manually define the set of corresponding 3D points $X$ on the  template for these semantic 2D keypoints. Given an input image with 2D annotations $\{x_i\}$ we can leverage these for learning. We do so by adding an objective  that ensures the projection of the corresponding 3D keypoints under the predicted camera pose $\pi$, after articulation, is consistent with the available 2D annotations. We denote $\mathcal{I}$ as indices of the visible keypoints to formalize the objective as:
\begin{align}
    L_{kp} = \sum_{i \in \mathcal{I}} \norm{x_i- \pi(\delta(X_i))}
\end{align}
In scenarios where such supervision is available, we observe that our approach enables us to easily leverage it for learning. While we later empirically examine such scenarios and highlight consistent benefits of allowing articulation in these, 
{\bf all visualizations in the paper are in a keypoint-free setting where this additional loss is not used}.

\vspace{2mm}
\noindent \textbf{Implementation Details.} We use a ResNet18 ~\cite{he2016deep} based encoder and a convolutional decoder to implement the per-pixel CSM predictor $f_{\theta}$ and another instantiation of ResNet18 based encoder for the deformation and camera predictor $g_{\theta'}$. We describe these in more detail in the supplemental and links to code are available on the webpage.

\section{Experiments}
Our approach allows us to: a) learn a  CSM prediction indicating mapping from each pixel to corresponding 3D point on template shape, and b) infer the articulation and pose that transforms the template to the image frame. We present experiments that evaluate both these aspects, and we empirically show that: a) allowing for articulation helps learn accurate CSM prediction (\secref{csmeval}), and b) we learn meaningful articulation, and that enforcing consistency with CSM is crucial for this learning  (\secref{arteval}).

\subsection{Datasets and Inputs}
\noindent
We create our dataset out of existing datasets -- CUB-200-2011~\cite{wah2011caltech},  PASCAL ~\cite{everingham2015pascal}, and Imagenet ~\cite{deng2009imagenet}, which we divide into
two sets by animal species. The first, ({\bf Set 1}) are birds, cows, horses and sheep, on which we report quantitative results. To demonstrate generality, we also have ({\bf Set 2}), or other animals on which we show qualitative results. Animals in Set 1 have keypoints available, which enable both quantitative results and experiments that test our model in the presence of keypoints. Animals in Set 2 do not have keypoints, and we show qualitative results. Throughout, we follow the underlying dataset's training and testing splits to ensure meaningful results.

\vspace{1mm}
\noindent {\bf Birds.} We use the CUB-200-2011 dataset for training and testing on birds (using the standard splits). It comprises 6000 images across 200 species, as well as foreground mask annotations (used for training), and annotated keypoints (used for evaluation, and optionally in training).

\begin{figure*}[h]
    \centering
    \includegraphics[width=\textwidth]{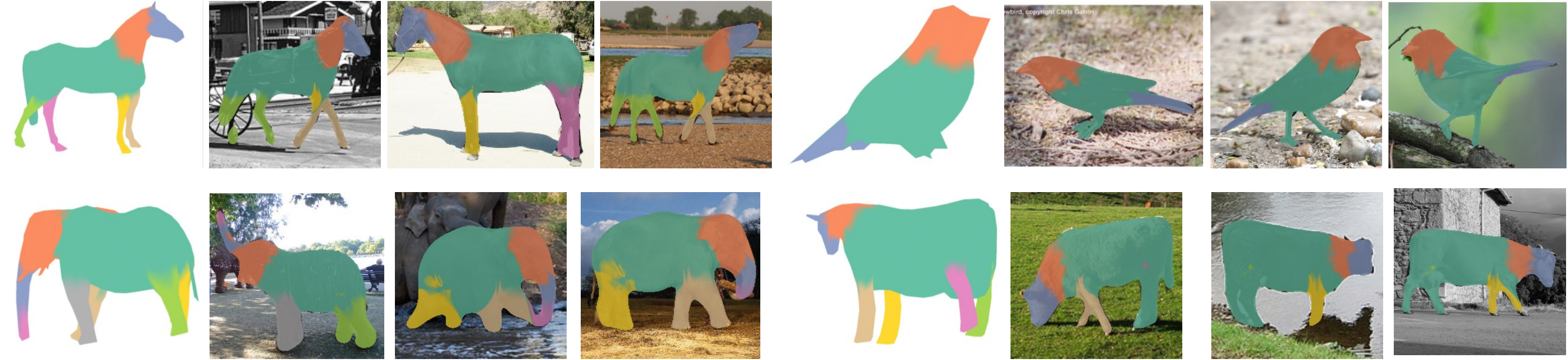}
    \caption{\textbf{Induced part labeling.} Our CSM inference allows inducing pixel-wise semantic part predictions. We visualize the parts of the template shape in the 1\textsuperscript{st} and 5\textsuperscript{th} columns, and the corresponding induced labels on the images via  corresponding 3D point. 
    }
    \figlabel{acsmparttransfer}
\end{figure*}

\vspace{1mm}
\noindent {\bf Set 1 Quadrupeds (Cows, Horses, Sheep).} We combine images from PASCAL VOC and  Imagenet. We use the VOC masks and masks on Imagenet produced from a COCO trained Mask RCNN model. When we report experiments that additionally leverage keypoints during training for these classes, they only use this supervision on the VOC training subset of images (and are therefore only `semi-supervised' in terms of keypoint annotations). 

\vspace{1mm}
\noindent {\bf Set 2 Quadrupeds (Hippos, Rhinos, Kangaroos, etc.).} We use images from Imagenet. In order to obtain masks for these animals, we annotate coarse masks for around  300 images per category, and then train a Mask RCNN by combining all these annotations into a single class, thus predicting segmentations for a generic `quadruped' category.

\vspace{1mm}
\noindent {\bf Filtering.} Throughout, we filter images with one large untruncated and largely unoccluded animal (\ie, some grass is fine).

\vspace{1mm}
\noindent {\bf Template Shapes.} We download models for all categories from~\cite{free3d}. We partition the quadrupeds to have 7 parts corresponding to torso, 4 legs, head, and a neck (see \figref{smooth_alpha} for examples of some of these). For the elephant model, we additionally mark two more parts for the trunk without a neck. Our birds model has 3 parts (head, torso, tail).

% as this dataset also has similar available annotations across these \ie segmentation masks and visible keypoints. 
\begin{figure}[h]
    \centering
    \includegraphics[width=0.47\textwidth]{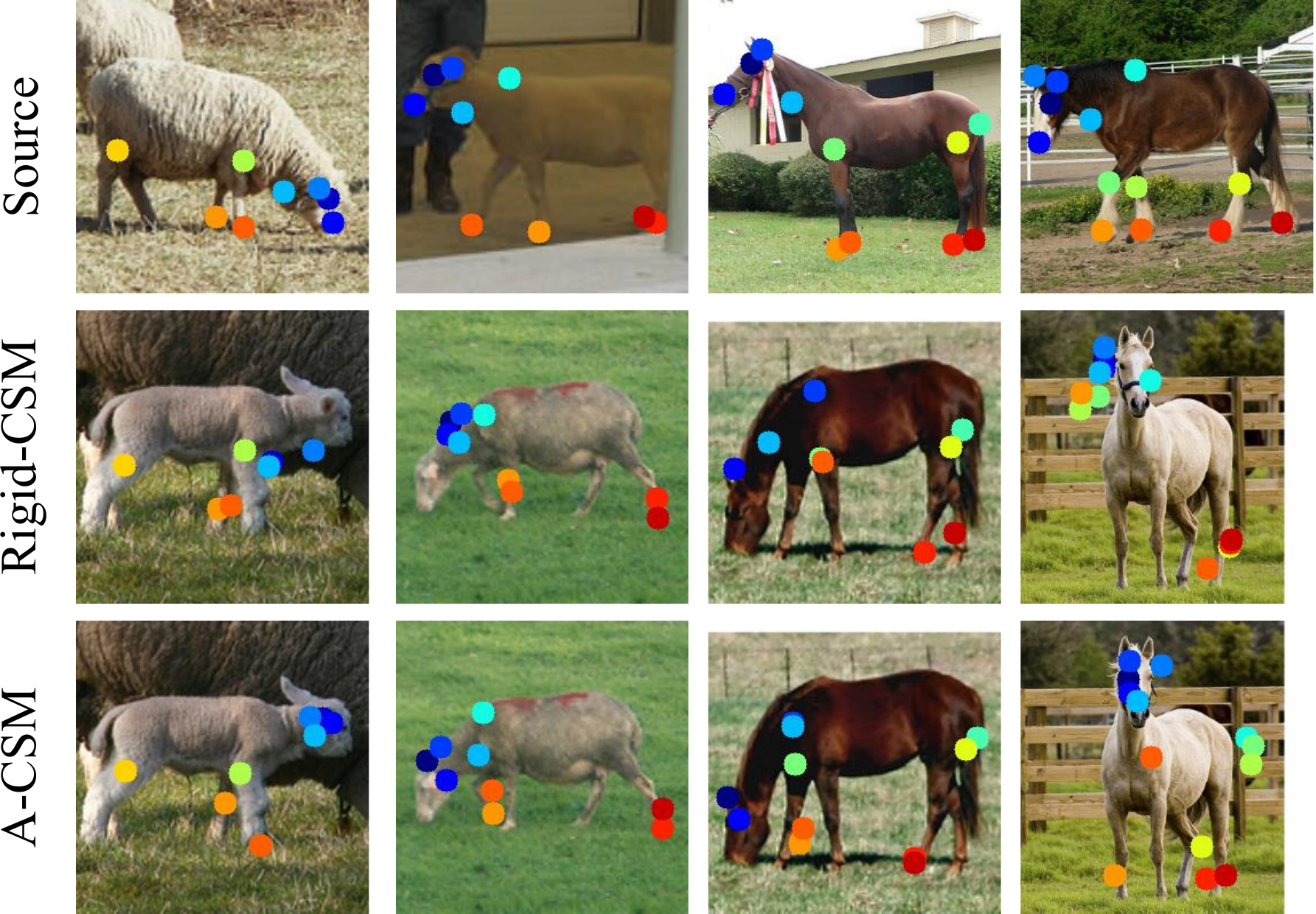}
    \caption{\textbf{Visualizing Keypoint Transfer.} We transfer keypoints from the `source' image to target image. Keypoint Transfer comparison between Rigid-CSM ~\cite{kulkarni2019canonical} and A-CSM (Ours). We see that the inferred correspondences as a result of modeling articulation are more accurate, for example note the keypoint transfers for the head of the sheep and horse.}
    \figlabel{kptransfer}
\end{figure}
\begin{table*}[!t]
\setlength{\tabcolsep}{0.02em}
\renewcommand{\arraystretch}{1}
\centering
  \scalebox{0.75}{
\begin{tabular}{lrl@{\hskip 0.05em}rl@{\hskip 0.05em}rl@{\hskip 0.05em}rl@{\hskip 0.05em}rl}
\addpic{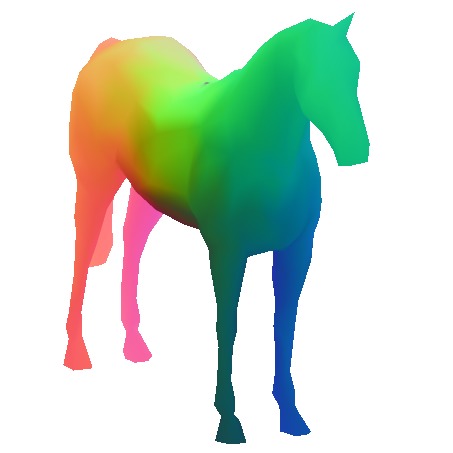} &
\addpic{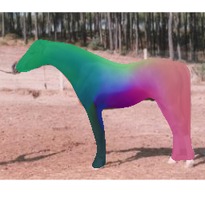} &
\addpichalf{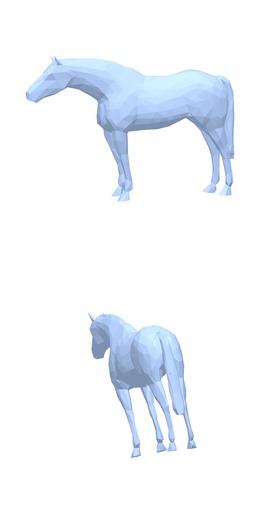} &
\addpic{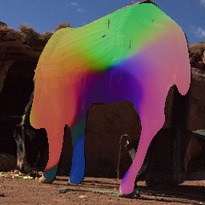} &
\addpichalf{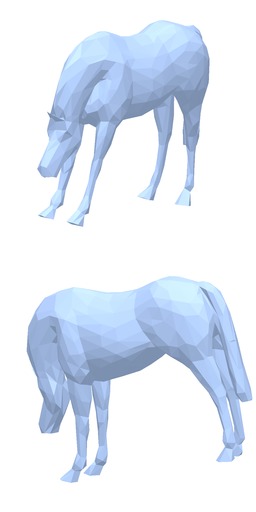} &
\addpic{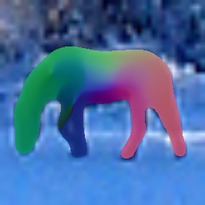} &
\addpichalf{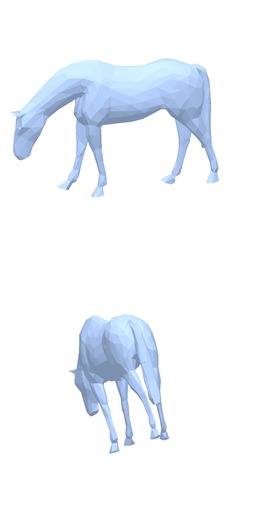} &  
\addpic{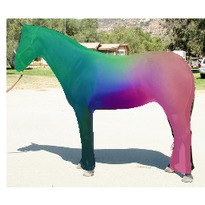} &
\addpichalf{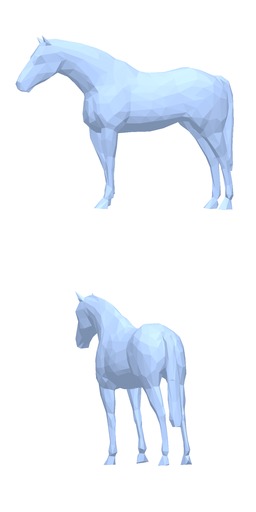} &
\addpic{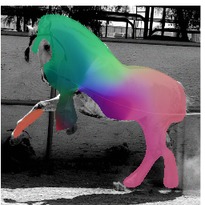} &
\addpichalf{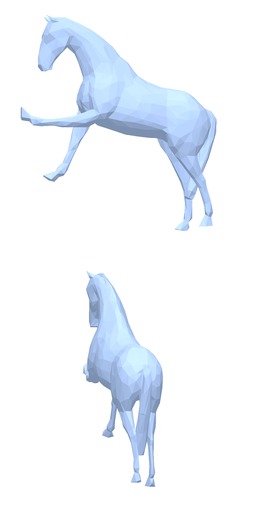} \\
% \addpic{figures/main_paper/horse_8parts/33/ind_33_uv_overlay.jpg} &
% \addpichalf{figures/main_paper/horse_8parts/33/ind_33_pose.jpg} \\  
%  & \addpic{figures/main_paper/horse_8parts/30/ind_30_uv_overlay.jpg} &
% \addpichalf{figures/main_paper/horse_8parts/30/ind_30_pose.jpg} &
% \addpic{figures/main_paper/horse_8parts/16/ind_16_uv_overlay.jpg} &
% \addpichalf{figures/main_paper/horse_8parts/16/ind_16_pose.jpg}\\
\midrule
\addpic{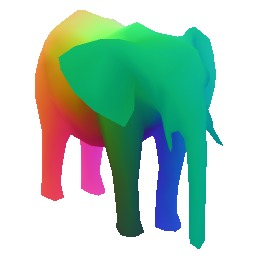} &
\addpic{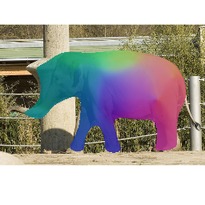} &
\addpichalf{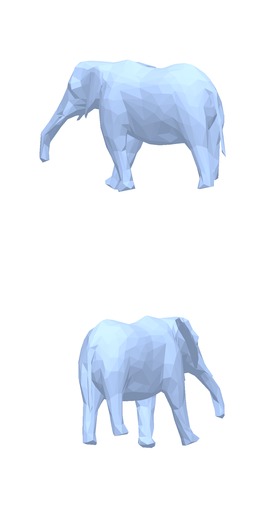} &
\addpic{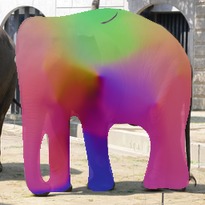} &
\addpichalf{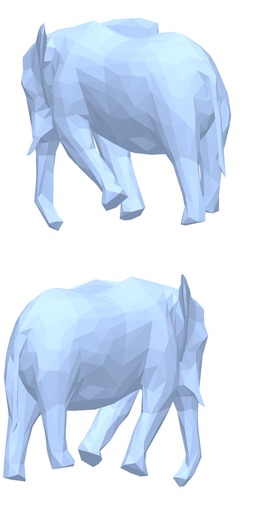} &
\addpic{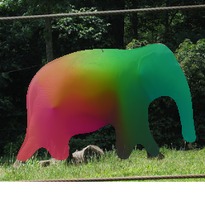} &
\addpichalf{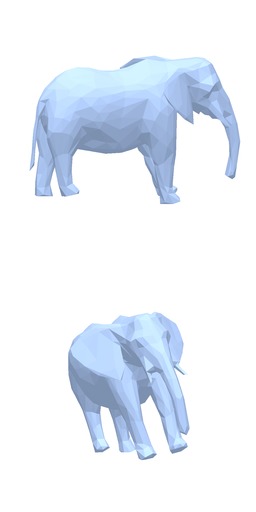} &
\addpic{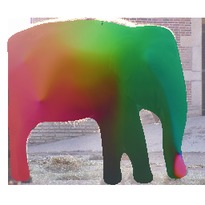} &
\addpichalf{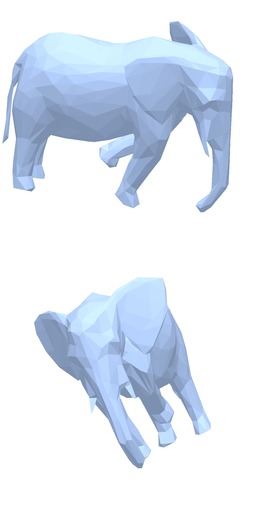} &
% \addpic{figures/main_paper/elephant/114/ind_114_uv_overlay.jpg} &
% \addpichalf{figures/main_paper/elephant/114/ind_114_pose.jpg} &
\addpic{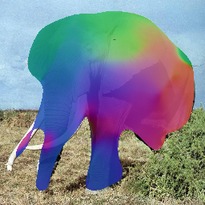} &
\addpichalf{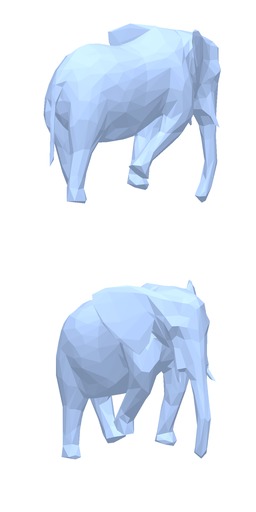} \\
% \addpic{figures/main_paper/elephant/85/ind_85_uv_overlay.jpg} &
% \addpichalf{figures/main_paper/elephant/85/ind_85_pose.jpg} &
% \addpic{figures/main_paper/elephant/23/ind_23_uv_overlay.jpg} &
% \addpichalf{figures/main_paper/elephant/23/ind_23_pose.jpg} &
% \addpic{figures/main_paper/elephant/123/ind_123_uv_overlay.jpg} &
% \addpichalf{figures/main_paper/elephant/123/ind_123_pose.jpg}
\midrule
\addpic{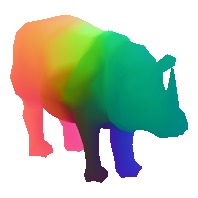} &
\addpic{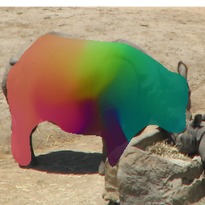} &
\addpichalf{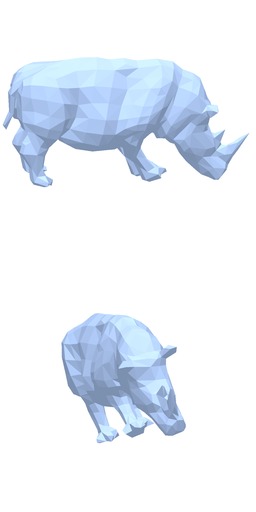} &
\addpic{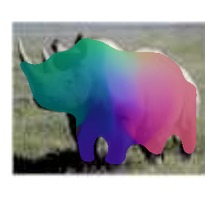} &
\addpichalf{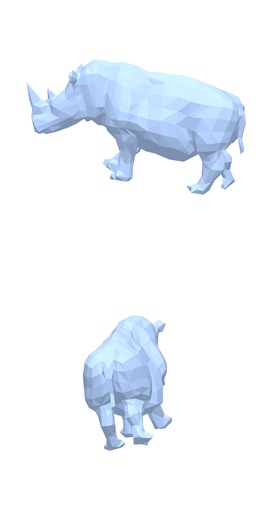} &
\addpic{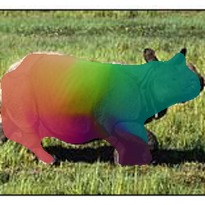} &
\addpichalf{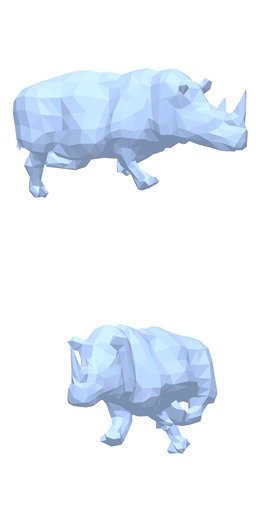} &
\addpic{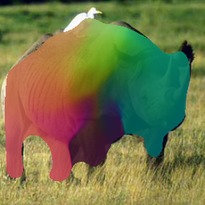} &
\addpichalf{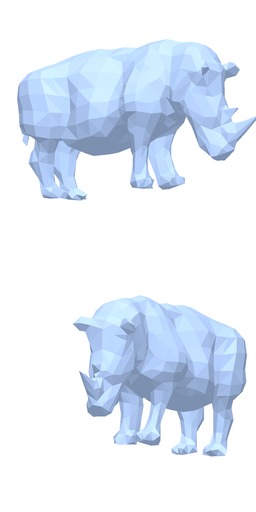} &
\addpic{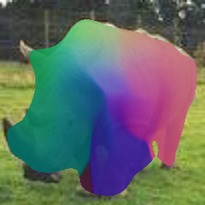} &
\addpichalf{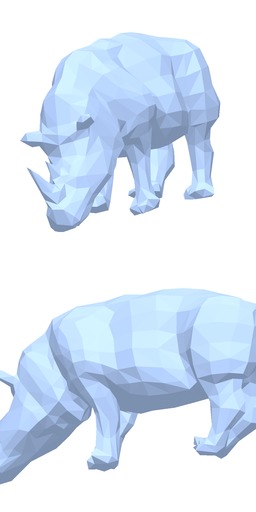}\\
% \addpic{figures/main_paper/rhino/134/ind_134_uv_overlay.jpg} &
% \addpichalf{figures/main_paper/rhino/134/ind_134_pose.jpg} &
\midrule
\addpic{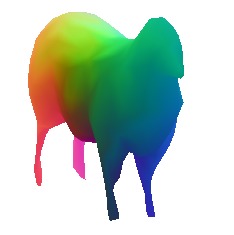} &
\addpic{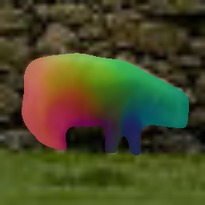} &
\addpichalf{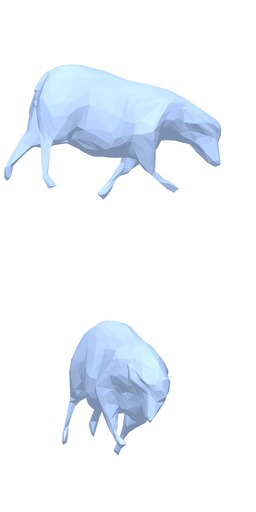} &
\addpic{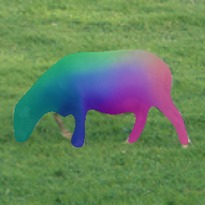} &
\addpichalf{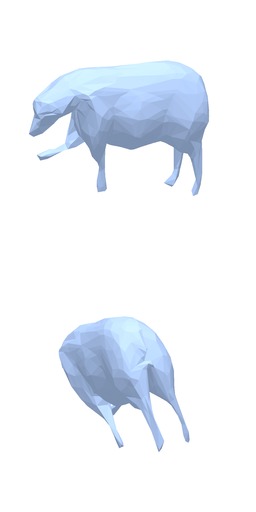} & 
 \addpic{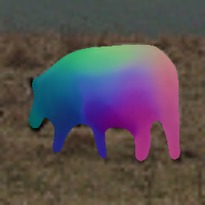} &
\addpichalf{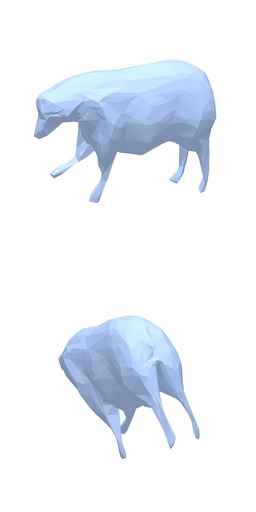} &
\addpic{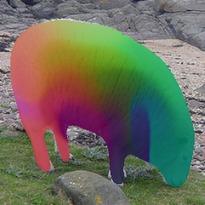} &
\addpichalf{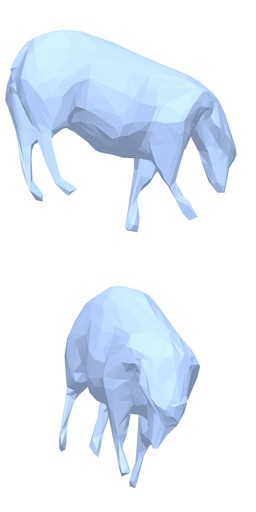} &
\addpic{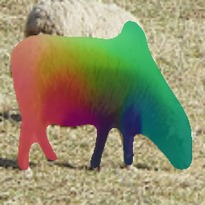} &
\addpichalf{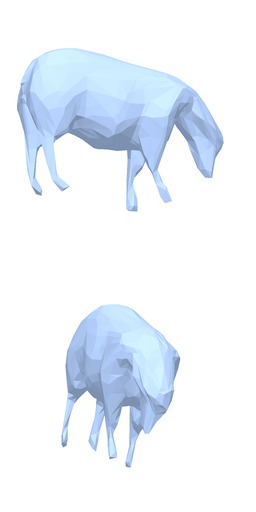} \\
\midrule
\addpic{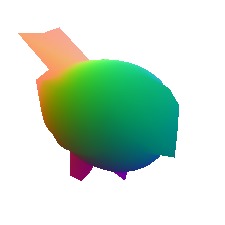} &
\addpic{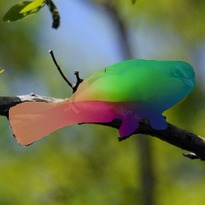} &
\addpichalf{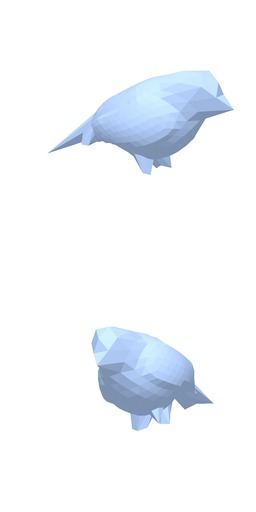} &
 \addpic{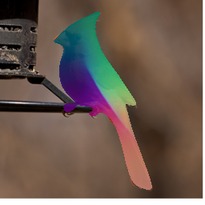} &
\addpichalf{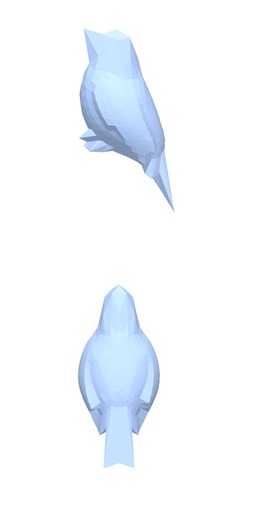} &
\addpic{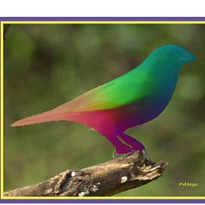} &
\addpichalf{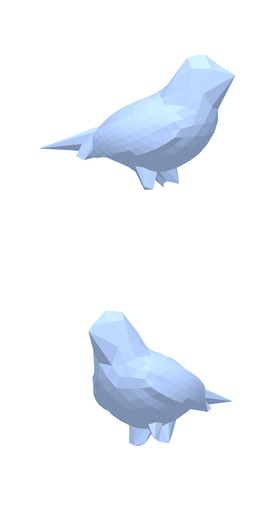}  &
  \addpic{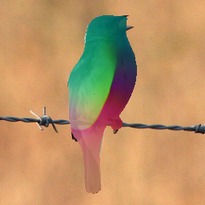} &
\addpichalf{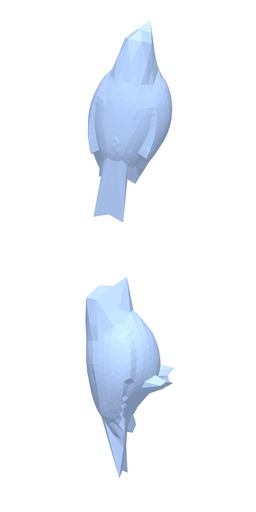} &
\addpic{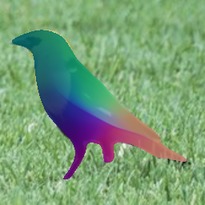} &
\addpichalf{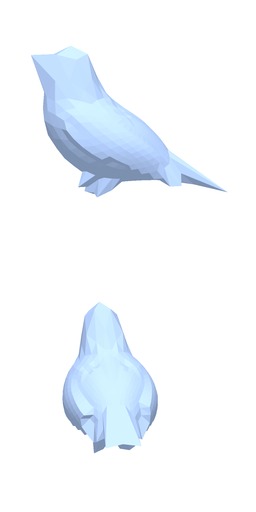} \\
\midrule
\addpic{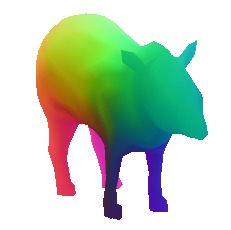} &
\addpic{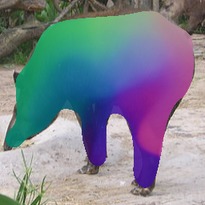} &
\addpichalf{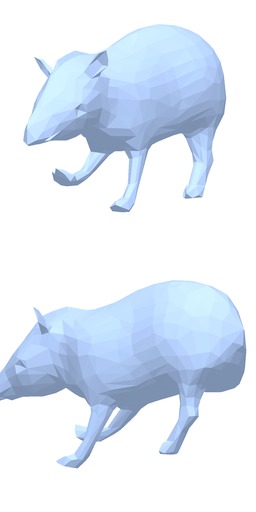} &
\addpic{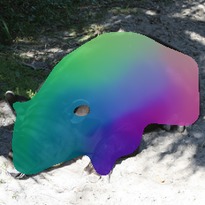} &
\addpichalf{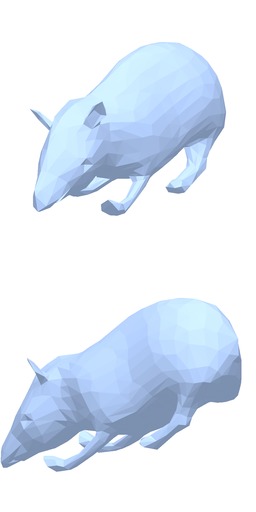} &
\addpic{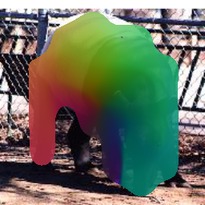} &
\addpichalf{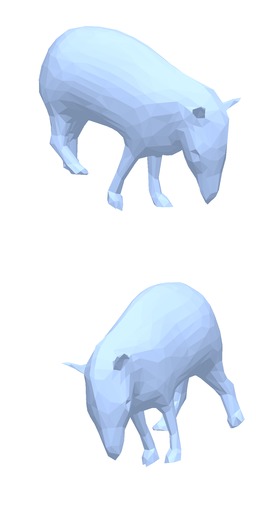} &
\addpic{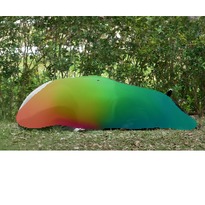} &
\addpichalf{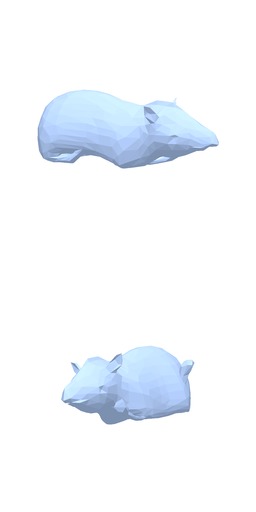} &
\addpic{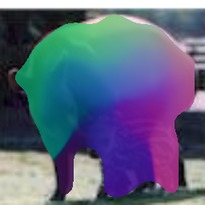} &
\addpichalf{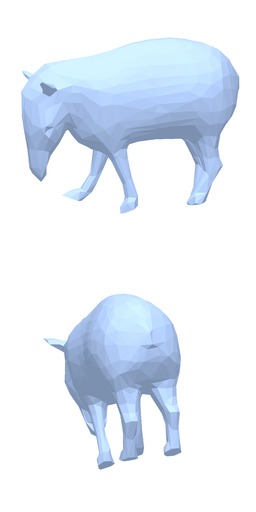} \\
\midrule
\addpic{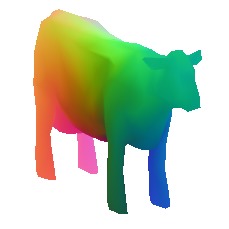} &
\addpic{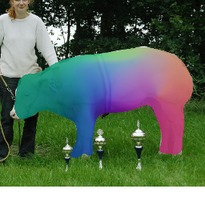} &
\addpichalf{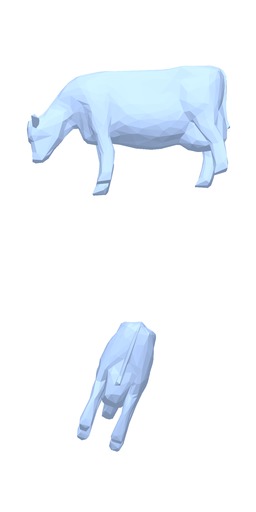} &
\addpic{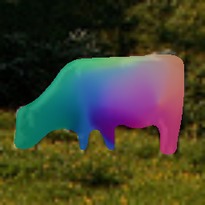} &
\addpichalf{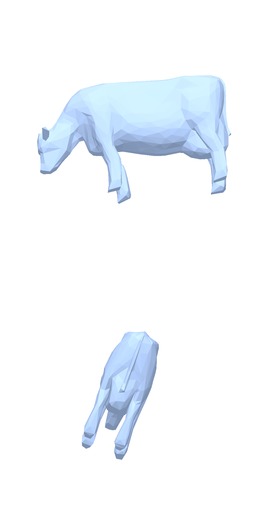} &
\addpic{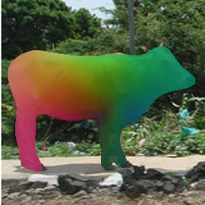} &
\addpichalf{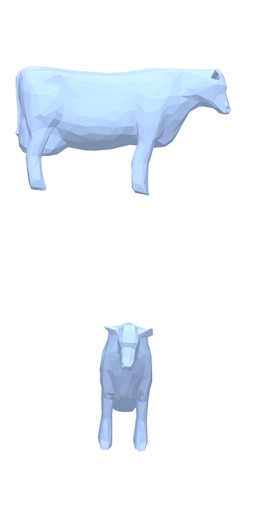} &
\addpic{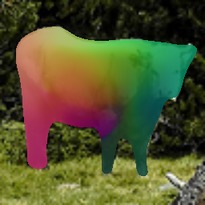} &
\addpichalf{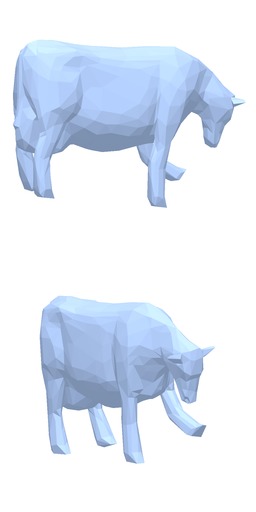} &
\addpic{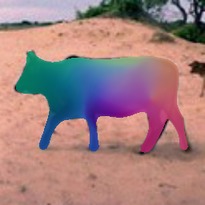} &
\addpichalf{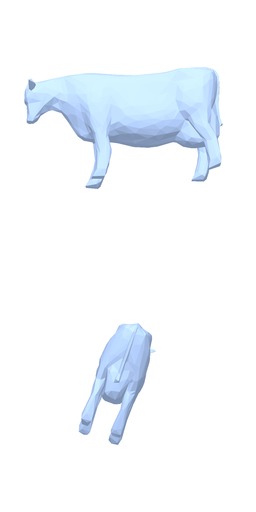} \\
\midrule
\addpic{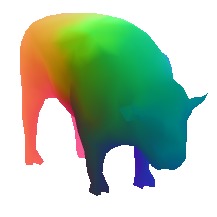} &
\addpic{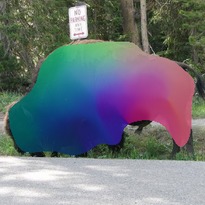} &
\addpichalf{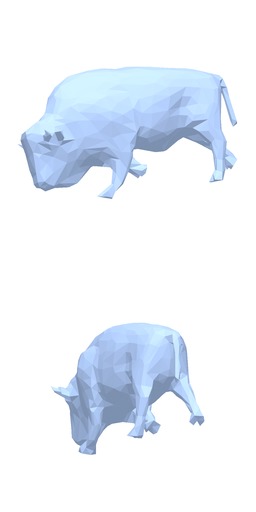} &
\addpic{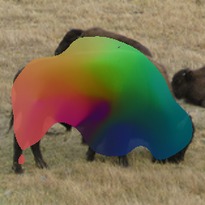} &
\addpichalf{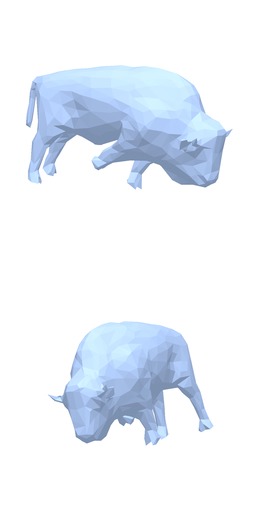} &
\addpic{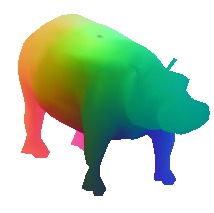} &
 & 
\addpic{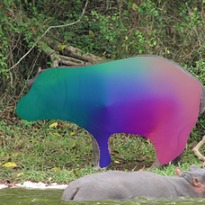} &
\addpichalf{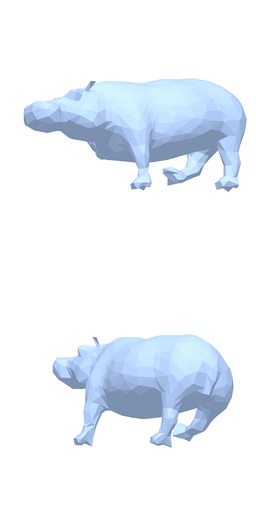} &
\addpic{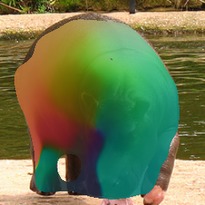} &
\addpichalf{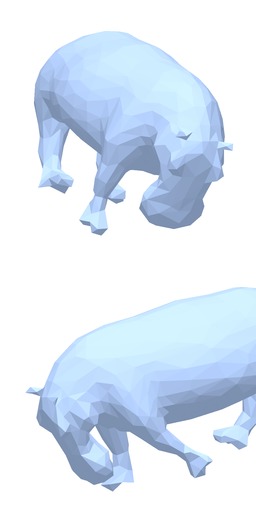} \\ 
\midrule
\addpic{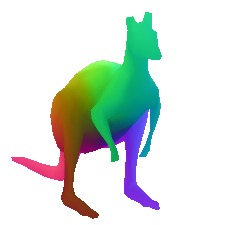} &
\addpic{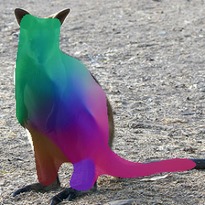} &
\addpichalf{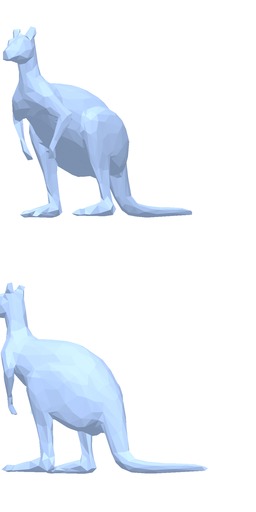} &
\addpic{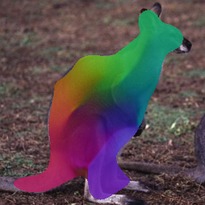} &
\addpichalf{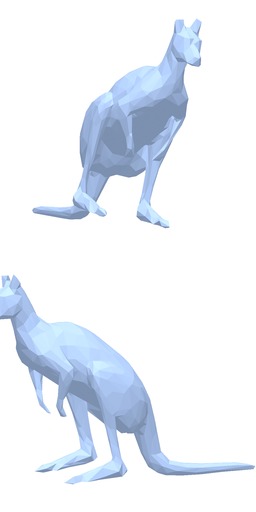} &
\addpic{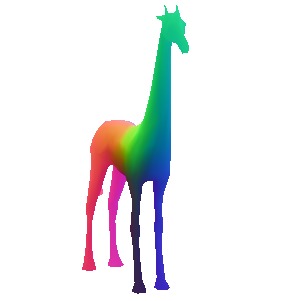} & &
\addpic{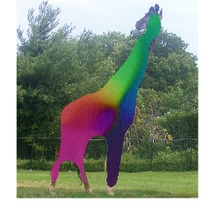} &
\addpichalf{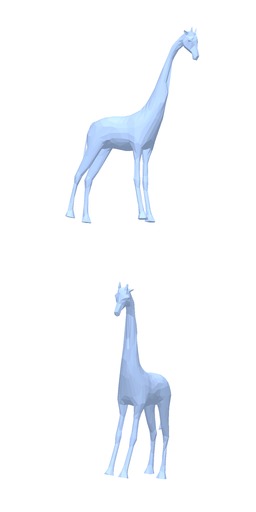} &
\addpic{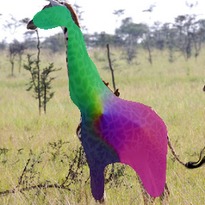} &
\addpichalf{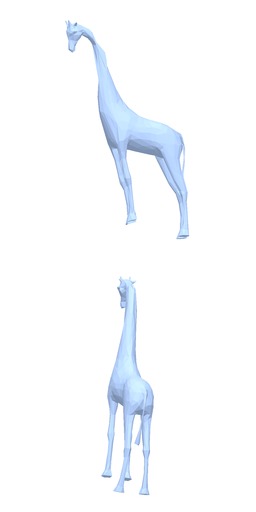}\\
\bottomrule \\
\end{tabular}
}
\vspace{-4mm}
\captionof{figure}{
\textbf{Sample Results.} We demonstrate our approach to learn the CSM mapping and articulation over a wide variety of non-rigid objects. The figure depicts: a) category-level the template shape on the left, b) per-image CSM prediction where colors indicate correspondence, and c) the predicted articulated shape from camera and a novel view. 
%The figure show articulations of template shape for every input image along side the CSM prediction for the foreground pixels. We observe consistent CSM predictions for various functional regions of the object. For instance, we can see the head of all the quadrupeds is greenish in color which accurately represents its mapping to green region on the template shape shown in the right-most column. We show results over 11 categories with a wide variety of articulations, but in certain cases we observe the model compensating by articulating excessively while sometimes under articulating.
}
\figlabel{qual-results-detection}
\end{table*}

\subsection{Evaluating CSM via Correspondence Transfer}
\seclabel{csmeval}
The predicted CSMs represent a per-pixel correspondence to the 3D template shape. Unfortunately, directly evaluating these requires dense annotation which is difficult to acquire, but we note that this prediction also allows one to infer dense correspondence across images. Therefore, we can follow the evaluation protocol typically used for measuring image to image correspondence quality~\cite{zhou2016learning,zhou2015flowweb}, and indirectly evaluate the learned mappings by measuring the accuracy for transferring annotated keypoints from a source to a target image as shown in \figref{kptransfer}. 

\vspace{1mm}
\noindent \textbf{Keypoint Transfer using CSM Prediction.} Given a source and a target image, we want to transfer an annotated keypoint from the source to the targert using the predicted pixelwise mapping. Intuitively, given a query source pixel, we can recover its corresponding 3D point on the template using the predicted CSM, and can then search over the target image for the pixel that is predicted to map the closest (described formally in the supplemental). Given some keypoint annotations on one image, we can therefore predict corresponding points on another.

\vspace{1mm}
\noindent \textbf{Evaluation Metric.} We use the `Percentage of Correct Keypoint Transfers' (PCK-Transfer) metric to indirectly evaluate the learned CSM mappings. Given several source-target image pairs, we transfer annotated keypoints from the source to the target, and label a transfer as `correct' if the predicted location is within $0.1\times~\max(w,h)$ distance of the ground-truth location. We report our performance over 10K source-target pairs

\vspace{1mm}
\noindent \textbf{Baselines.} We report comparisons against two alternate approaches that leverage similar form of supervision. First, we compare against Rigid-CSM ~\cite{kulkarni2019canonical} which learns similar pixel to surface mappings, but without allowing model articulation. The implementation of this baseline simply corresponds to using our training approach, but without any the articulation $\delta$.
We also compare against the Dense Equivariance (DE)~\cite{thewlis2017unsupervised} approach that learns self-supervised mappings from pixels to an implicit (and non-geometric) space.

\begin{table}[t!]
\caption{{\bf PCK-Transfer for Evaluating CSM Prediction}. We evaluate the transfer of keypoints from a source and target image, and report the transfer accuracy as PCK transfer as described in \secref{csmeval}. Higher is better
% The best numbers are in \textbf{bold}.7777
}
\tablelabel{pcktransfer}
\setlength{\tabcolsep}{4pt} 
\centering 
% \small{
% \scalebox{1.0}{
\resizebox{
        \width
      }{!}{
\begin{tabular}{@{}ll|rrrr@{}} \toprule
% \multicolumn{2}{c}{Item} \\ \cmidrule(r){1-2}
Supv & Method & Birds & Horses   & Cows &Sheep\\ \midrule
\multirow{2}{1cm}{KP + Mask} & Rigid-CSM ~\cite{kulkarni2019canonical}  & 45.8 & 42.1 & 28.5 & 31.5 \\ 
 & A-CSM (ours)      & 51.0   & 44.6 & 29.2 & 39.0   \\ \midrule
\multirow{2}{1cm}{Mask} & Rigid-CSM ~\cite{kulkarni2019canonical} & 36.4 & 31.2 & 26.3 & 24.7 \\
      & Dense-Equi ~\cite{thewlis2017unsupervised} & 33.5 & 23.3 & 20.9 & 19.6 \\
       &  A-CSM (ours)      & 42.6 & 32.9 & 26.3 & 28.6 \\
 \bottomrule
\end{tabular}
}
% }

\end{table}

\vspace{1mm}
\noindent \textbf{Results.} We report the empirical results obtained under two settings: with, and without keypoint supervision for learning in \tableref{pcktransfer}. We find that across both these settings, our approach of learning pixel to surface mappings using articulation-aware geometric consistency improves over learning using articulation-agnostic consistency.  We also find that our geometry-aware approach performs better than learning equivariant embeddings using synthetic transforms. We visualize keypoint transfer results in \figref{kptransfer} and observe accurate transfers despite different articulation \eg for the horse head, we can accurately transfer the keypoints despite it being bent in the target and not in the source. The Rigid-CSM~\cite{kulkarni2019canonical} baseline however, does not do so successfully. We also visualize the induced part labeling by transferring part labels from 3D models to image pixels shown in \figref{acsmparttransfer}.

\subsection{Articulation Evaluation via Keypoint Reprojection.}
\seclabel{arteval}
\noindent
Towards analyzing the fidelity of the learned articulation (and pose), we observe that under accurate predictions, annotated 2D keypoints in images should match re-projection of manually defined 3D keypoints on the template. We therefore measure whether the 3D keypoints on the articulated template, when reprojected back with the predicted camera pose, match the 2D annotations. Using this metric, we address: a) does allowing articulation help accuracy? and b) is joint training with CSM consistency helpful?

\vspace{1mm}
\noindent \textbf{Evaluation Metrics.} We again use the `Percentage of Correct Keypoints' (PCK) metric to evaluate the accuracy of 3D keypoints of template when articulated and reprojected. For each test image with available 2D keypoint keypoint annotations, we obtain reprojections of 3D points, and label a reprojection correct if the predicted location is within $0.1 \times ~\max(w,h)$ distance of the ground-truth. Note that unlike `PCK-Transfer', this evaluation is done per-image.

\begin{table}[t!]
\caption{{\bf Articulation Evaluation}. We compute PCK under reprojection of manually annotated keypoints on the mesh as described in \secref{arteval}. Higher is better. 
% The best numbers are in \textbf{bold}.
}
% }
\tablelabel{pck}
\setlength{\tabcolsep}{4pt} 
\centering 
% \small{
% \scalebox{1.0}{
\resizebox{
        \width
      }{!}{
\begin{tabular}{@{}ll|rrrr@{}} \toprule
% \multicolumn{2}{c}{Item} \\ \cmidrule(r){1-2}
Supv & Method & Birds & Horses   & Cows &Sheep\\ \midrule
\multirow{2}{1cm}{KP + Mask} & Rigid-CSM ~\cite{kulkarni2019canonical}  & 68.5 & 46.4 & 52.6 & 47.9 \\
 & A-CSM (ours) & 72.4 & 57.3 & 56.8 & 57.4 \\  \midrule
\multirow{2}{1cm}{Mask}     & Rigid-CSM ~\cite{kulkarni2019canonical} & 50.9 & 49.7 & 37.4 & 36.4\\
  &   A-CSM (ours) & 46.8 & 54.2 & 43.8 & 42.5 \\
 \bottomrule
\end{tabular}
}

\end{table}

\vspace{1mm}
\noindent \textbf{Do we learn meaningful articulation?} 
We report the keypoint reprojection accuracy across classes under settings with different forms of supervision in \tableref{pck}. We compare against the alternate approach of not modeling articulations, and observe that our approach yields more accurate predictions, thereby highlighting that we do learn meaningful articulation. One exception is for `birds' when training without keypoint supervision, but we find this to occur because of some ambiguities in defining the optimal 3D keypoint on the template, as `back', `wing' \etc and we found that our model simply learned a slightly different (but consistent) notion of pose, leading to suboptimal evaluation. We also show several qualitative results in \figref{qual-results-detection} and \figref{teaser} that depict articulations of the canonical mesh for various input images, and do observe that we can learn to articulate parts like moving legs, elephant trunk, animal heads \etc, and these results do clearly highlight that we can learn articulation using our approach.

\begin{table}[t!]
\caption{{\bf Effect of $L_{gcc}$ for Learning Articulation}. We report performance of our method, and compare it with a variant trained without the geometric cycle loss.
% We observe a significant drop in PCK performance w/o the GCC constraint in the setting without keypoints, while in the setting with keypoints the camera and articulations train well with the keypoint loss.  
% The best numbers are in \textbf{bold}.
}
\tablelabel{ablation}
\setlength{\tabcolsep}{4pt} 
\centering 
% \small{
% \scalebox{1.0}{
\resizebox{
        \width
      }{!}{

\begin{tabular}{@{}ll|rrrr@{}} \toprule
% \multicolumn{2}{c}{Item} \\ \cmidrule(r){1-2}
Supv & Method & Birds & Horses   & Cows &Sheep\\ \midrule
\multirow{2}{1cm}{KP + Mask} & A-CSM (ours)       & 72.4 & 57.3 & 56.8 & 57.4 \\
 & A-CSM  w/o GCC & 72.2 & 35.5 & 56.6 & 54.5 \\ \midrule
\multirow{2}{1cm}{Mask}    & A-CSM (ours)       & 46.8 & 54.2 & 43.8 & 42.5 \\
      & A-CSM  w/o GCC & 12.9 & 24.8 & 18.7 & 16.6 \\
 \bottomrule
\end{tabular}
}
% }

\end{table}

\vspace{1mm}
\noindent \textbf{Does consistency with CSM help learn articulation?}
The cornerstone of our approach is that we can obtain supervisory signal by enforcing consistency among predicted CSM, articulation, and pose. However, another source of signal for learning articulation (and pose) is the mask supervision. We therefore investigate whether this joint consistency is useful for learning, or whether just the mask supervision can suffice. We train a variant of our model `A-CSM w/o GCC' where we only learn the pose and articulation predictor $g$, without the cycle consistency loss. We report the results obtained under two supervision settings in \tableref{ablation}, and find that when keypoint supervision is available, using the consistency gives modest improvements. However, when keypoint supervision is not available, we observe that this consistency is critical for learning articulation (and pose), and that performance in settings without keypoint supervision drops significantly if not enforced.

\subsection{Learning from Imagenet}
\seclabel{imgnet}
\noindent
As our approach enables learning pixel to surface mappings and articulation without requiring keypoint supervision, we can learn these from a category-level image collection \eg ImageNet, using automatically obtained segmentation masks. 
We used our `quadruped' trained Mask-RCNN to obtain (noisy) segmentation masks per instance.  
We then use our approach to learn articulation and canonical surface mapping for these classes. We show some results in \figref{teaser} and \figref{qual-results-detection}, where all classes except (birds, horse, sheep, cow) were trained using only ImageNet images. We observe that even under this setting with limited and noisy supervision, our approach enables us to learn meaningful articulation and consistent CSM prediction.

\section{Discussion}
\noindent
We presented an approach to jointly learn prediction of canonical surface mappings and articulation, without direct supervision, by instead enforcing consistency among the predictions. While enabling articulations allowed us to go beyond explaining pixelwise predictions via reprojections of a rigid template, the class of transformations allowed may still be restrictive in case of intrinsic shape variation. An even more challenging scenario where our approach is not directly applicable is for categories where a template is not well-defined \eg chairs, and future attempts could investigate enabling learning over these. Finally, while our focus was to demonstrate results in setting without direct supervision, our techniques may also be applicable in scenarios where large-scale annotation is available, and can serve as further regularization or a mechanism to include even more unlabelled data for learning. 
\\
% \vspace{-2mm}% \vspace{2mm}
\noindent
% \vspace{-2mm}
{
{\bf Acknowledgements.} We would like to thank the members of the Fouhey AI lab (FAIL), CMU Visual Robot Learning lab and anonymous reviewers for helpful discussions and feedback. We also thank Richard Higgins for his help with varied quadruped category suggestions and annotating 3D models}
%\clearpage
{\small
\bibliographystyle{ieee_fullname}
\bibliography{egbib}

\begin{thebibliography}{10}\itemsep=-1pt

\bibitem{free3d}
Free3d.com.
\newblock \url{http://www.free3d.com}.

\bibitem{alp2018densepose}
R{\i}za Alp~G{\"u}ler, Natalia Neverova, and Iasonas Kokkinos.
\newblock Densepose: Dense human pose estimation in the wild.
\newblock In {\em CVPR}, 2018.

\bibitem{BlanzVetter}
Volker Blanz and Thomas Vetter.
\newblock A morphable model for the synthesis of 3d faces.
\newblock In {\em SIGGRAPH}, 1999.

\bibitem{bogo2016keep}
Federica Bogo, Angjoo Kanazawa, Christoph Lassner, Peter Gehler, Javier Romero,
  and Michael~J Black.
\newblock Keep it smpl: Automatic estimation of 3d human pose and shape from a
  single image.
\newblock In {\em ECCV}. Springer, 2016.

\bibitem{choy2016universal}
Christopher~B Choy, JunYoung Gwak, Silvio Savarese, and Manmohan Chandraker.
\newblock Universal correspondence network.
\newblock In {\em NeurIPS}, 2016.

\bibitem{LocalChapterEvents:ItalChap:ItalianChapConf2008:129-136}
Paolo Cignoni, Marco Callieri, Massimiliano Corsini, Matteo Dellepiane, Fabio
  Ganovelli, and Guido Ranzuglia.
\newblock {MeshLab: an Open-Source Mesh Processing Tool}.
\newblock In Vittorio Scarano, Rosario~De Chiara, and Ugo Erra, editors, {\em
  Eurographics Italian Chapter Conference}. The Eurographics Association, 2008.

\bibitem{deng2009imagenet}
Jia Deng, Wei Dong, Richard Socher, Li-Jia Li, Kai Li, and Li Fei-Fei.
\newblock Imagenet: A large-scale hierarchical image database.
\newblock In {\em CVPR}, 2009.

\bibitem{everingham2015pascal}
Mark Everingham, SM~Ali Eslami, Luc Van~Gool, Christopher~KI Williams, John
  Winn, and Andrew Zisserman.
\newblock The pascal visual object classes challenge: A retrospective.
\newblock {\em IJCV}, 2015.

\bibitem{fan2017point}
Haoqiang Fan, Hao Su, and Leonidas~J Guibas.
\newblock A point set generation network for 3d object reconstruction from a
  single image.
\newblock In {\em CVPR}, 2017.

\bibitem{florence2018dense}
Peter~R Florence, Lucas Manuelli, and Russ Tedrake.
\newblock Dense object nets: Learning dense visual object descriptors by and
  for robotic manipulation.
\newblock {\em CoRL}, 2018.

\bibitem{garland1997surface}
Michael Garland and Paul~S Heckbert.
\newblock Surface simplification using quadric error metrics.
\newblock In {\em SIGGRAPH}. ACM Press/Addison-Wesley Publishing Co., 1997.

\bibitem{girdhar2016learning}
Rohit Girdhar, David~F Fouhey, Mikel Rodriguez, and Abhinav Gupta.
\newblock Learning a predictable and generative vector representation for
  objects.
\newblock In {\em ECCV}. Springer, 2016.

\bibitem{he2017mask}
Kaiming He, Georgia Gkioxari, Piotr Doll{\'a}r, and Ross Girshick.
\newblock Mask r-cnn.
\newblock In {\em ICCV}, 2017.

\bibitem{he2016deep}
Kaiming He, Xiangyu Zhang, Shaoqing Ren, and Jian Sun.
\newblock Deep residual learning for image recognition.
\newblock In {\em CVPR}, 2016.

\bibitem{huttenlocher1990recognizing}
Daniel~P Huttenlocher and Shimon Ullman.
\newblock Recognizing solid objects by alignment with an image.
\newblock {\em IJCV}, 1990.

\bibitem{insafutdinov2018unsupervised}
Eldar Insafutdinov and Alexey Dosovitskiy.
\newblock Unsupervised learning of shape and pose with differentiable point
  clouds.
\newblock In {\em NeurIPS}, 2018.

\bibitem{hmrKanazawa17}
Angjoo Kanazawa, Michael~J. Black, David~W. Jacobs, and Jitendra Malik.
\newblock End-to-end recovery of human shape and pose.
\newblock In {\em CVPR}, 2018.

\bibitem{kanazawa2016learning}
Angjoo Kanazawa, Shahar Kovalsky, Ronen Basri, and David Jacobs.
\newblock Learning 3d deformation of animals from 2d images.
\newblock In {\em Eurographics}, volume~35, pages 365--374. Wiley Online
  Library, 2016.

\bibitem{cmrKanazawa18}
Angjoo Kanazawa, Shubham Tulsiani, Alexei~A. Efros, and Jitendra Malik.
\newblock Learning category-specific mesh reconstruction from image
  collections.
\newblock In {\em ECCV}, 2018.

\bibitem{kar2015category}
Abhishek Kar, Shubham Tulsiani, Joao Carreira, and Jitendra Malik.
\newblock Category-specific object reconstruction from a single image.
\newblock In {\em CVPR}, 2015.

\bibitem{kingma2014adam}
Diederik~P Kingma and Jimmy Ba.
\newblock Adam: A method for stochastic optimization.
\newblock {\em arXiv preprint arXiv:1412.6980}, 2014.

\bibitem{kulkarni2019canonical}
Nilesh Kulkarni, Abhinav Gupta, and Shubham Tulsiani.
\newblock Canonical surface mapping via geometric cycle consistency.
\newblock In {\em ICCV}, 2019.

\bibitem{lewis2000pose}
John~P Lewis, Matt Cordner, and Nickson Fong.
\newblock Pose space deformation: a unified approach to shape interpolation and
  skeleton-driven deformation.
\newblock In {\em Proceedings of the 27th annual conference on Computer
  graphics and interactive techniques}, pages 165--172. ACM
  Press/Addison-Wesley Publishing Co., 2000.

\bibitem{lin2018learning}
Chen-Hsuan Lin, Chen Kong, and Simon Lucey.
\newblock Learning efficient point cloud generation for dense 3d object
  reconstruction.
\newblock In {\em AAAI}, 2018.

\bibitem{lin2014microsoft}
Tsung-Yi Lin, Michael Maire, Serge Belongie, James Hays, Pietro Perona, Deva
  Ramanan, Piotr Doll{\'a}r, and C~Lawrence Zitnick.
\newblock Microsoft coco: Common objects in context.
\newblock In {\em ECCV}, 2014.

\bibitem{SMPL}
Matthew Loper, Naureen Mahmood, Javier Romero, Gerard Pons-Moll, and Michael~J.
  Black.
\newblock {SMPL}: A skinned multi-person linear model.
\newblock {\em SIGGRAPH Asia}, 2015.

\bibitem{lowe2004sift}
G Lowe.
\newblock Sift-the scale invariant feature transform.
\newblock {\em IJCV}, 2004.

\bibitem{maron2017convolutional}
Haggai Maron, Meirav Galun, Noam Aigerman, Miri Trope, Nadav Dym, Ersin Yumer,
  Vladimir~G Kim, and Yaron Lipman.
\newblock Convolutional neural networks on surfaces via seamless toric covers.
\newblock 2017.

\bibitem{neverova2019slim}
Natalia Neverova, James Thewlis, Riza~Alp Guler, Iasonas Kokkinos, and Andrea
  Vedaldi.
\newblock Slim densepose: Thrifty learning from sparse annotations and motion
  cues.
\newblock In {\em CVPR}, 2019.

\bibitem{oberweger2015hands}
Markus Oberweger, Paul Wohlhart, and Vincent Lepetit.
\newblock Hands deep in deep learning for hand pose estimation.
\newblock {\em arXiv preprint arXiv:1502.06807}, 2015.

\bibitem{paszke2017automatic}
Adam Paszke, Sam Gross, Soumith Chintala, Gregory Chanan, Edward Yang, Zachary
  DeVito, Zeming Lin, Alban Desmaison, Luca Antiga, and Adam Lerer.
\newblock Automatic differentiation in {PyTorch}.
\newblock In {\em NIPS Autodiff Workshop}, 2017.

\bibitem{pavlakos2019expressive}
Georgios Pavlakos, Vasileios Choutas, Nima Ghorbani, Timo Bolkart, Ahmed~AA
  Osman, Dimitrios Tzionas, and Michael~J Black.
\newblock Expressive body capture: 3d hands, face, and body from a single
  image.
\newblock In {\em CVPR}, 2019.

\bibitem{pepik2012teaching}
Bojan Pepik, Michael Stark, Peter Gehler, and Bernt Schiele.
\newblock Teaching 3d geometry to deformable part models.
\newblock In {\em CVPR}, 2012.

\bibitem{sinha2017surfnet}
Ayan Sinha, Asim Unmesh, Qixing Huang, and Karthik Ramani.
\newblock Surfnet: Generating 3d shape surfaces using deep residual networks.
\newblock In {\em Proceedings of the IEEE conference on computer vision and
  pattern recognition}, pages 6040--6049, 2017.

\bibitem{su2015render}
Hao Su, Charles~R Qi, Yangyan Li, and Leonidas~J Guibas.
\newblock Render for cnn: Viewpoint estimation in images using cnns trained
  with rendered 3d model views.
\newblock In {\em ICCV}, 2015.

\bibitem{thewlis2017unsupervised}
James Thewlis, Hakan Bilen, and Andrea Vedaldi.
\newblock Unsupervised learning of object frames by dense equivariant image
  labelling.
\newblock In {\em NeurIPS}, 2017.

\bibitem{mvcTulsiani18}
Shubham Tulsiani, Alexei~A. Efros, and Jitendra Malik.
\newblock Multi-view consistency as supervisory signal for learning shape and
  pose prediction.
\newblock In {\em CVPR}, 2018.

\bibitem{tulsiani2015viewpoints}
Shubham Tulsiani and Jitendra Malik.
\newblock Viewpoints and keypoints.
\newblock In {\em CVPR}, 2015.

\bibitem{drcTulsiani17}
Shubham Tulsiani, Tinghui Zhou, Alexei~A. Efros, and Jitendra Malik.
\newblock Multi-view supervision for single-view reconstruction via
  differentiable ray consistency.
\newblock In {\em CVPR}, 2017.

\bibitem{wah2011caltech}
Catherine Wah, Steve Branson, Peter Welinder, Pietro Perona, and Serge
  Belongie.
\newblock The caltech-ucsd birds-200-2011 dataset.
\newblock 2011.

\bibitem{Wang_2019_CVPR}
He Wang, Srinath Sridhar, Jingwei Huang, Julien Valentin, Shuran Song, and
  Leonidas~J. Guibas.
\newblock Normalized object coordinate space for category-level 6d object pose
  and size estimation.
\newblock In {\em CVPR}, 2019.

\bibitem{xiang2019monocular}
Donglai Xiang, Hanbyul Joo, and Yaser Sheikh.
\newblock Monocular total capture: Posing face, body, and hands in the wild.
\newblock In {\em CVPR}, 2019.

\bibitem{xiang2014beyond}
Yu Xiang, Roozbeh Mottaghi, and Silvio Savarese.
\newblock Beyond pascal: A benchmark for 3d object detection in the wild.
\newblock In {\em WACV}, 2014.

\bibitem{xiang2017posecnn}
Yu Xiang, Tanner Schmidt, Venkatraman Narayanan, and Dieter Fox.
\newblock Posecnn: A convolutional neural network for 6d object pose estimation
  in cluttered scenes.
\newblock {\em RSS 2018}, 2017.

\bibitem{yan2016perspective}
Xinchen Yan, Jimei Yang, Ersin Yumer, Yijie Guo, and Honglak Lee.
\newblock Perspective transformer nets: Learning single-view 3d object
  reconstruction without 3d supervision.
\newblock In {\em NeurIPS}, 2016.

\bibitem{zhou2015flowweb}
Tinghui Zhou, Yong Jae~Lee, Stella~X Yu, and Alyosha~A Efros.
\newblock Flowweb: Joint image set alignment by weaving consistent, pixel-wise
  correspondences.
\newblock In {\em CVPR}, 2015.

\bibitem{zhou2016learning}
Tinghui Zhou, Philipp Kr{\"a}henb{\"u}hl, Mathieu Aubry, Qixing Huang, and
  Alexei~A. Efros.
\newblock Learning dense correspondence via 3d-guided cycle consistency.
\newblock In {\em CVPR}, 2016.

\bibitem{zhu2016face}
Xiangyu Zhu, Zhen Lei, Xiaoming Liu, Hailin Shi, and Stan~Z Li.
\newblock Face alignment across large poses: A 3d solution.
\newblock In {\em CVPR}, 2016.

\bibitem{zuffi2019three}
Silvia Zuffi, Angjoo Kanazawa, Tanja Berger-Wolf, and Michael~J Black.
\newblock Three-d safari: Learning to estimate zebra pose, shape, and texture
  from images" in the wild".
\newblock 2019.

\end{thebibliography}
}

%%%%%%%%% ABSTRACT

%%%%%%%%% BODY TEXT
% TODO
% \begin{itemize}
%     % \item Mesh Preliminaries
%     % \item Articulating models
%     % \item Barycentric Interpolation
%     % \item Visibility constraints
%     % \item Mask loss
%     % \item Keypoint transfer using CSM Prediction
%     % \item Network Details
%     \item Dump visualization
%     \item Create Video
% \end{itemize}
% {\normal }
\clearpage
\setcounter{section}{0}
\renewcommand{\thesection}{\Alph{section}}
\section{Constructing $\phi$ and $\delta$ }
\noindent
{\bf Parameterizing surface of a mesh.} The surface $S$ of a mesh is 2D manifold in 3D space hence we can construct a mapping $\phi : [0,1)^2 \xrightarrow{} S$. We deal with triangular meshes as they are the most general form of mesh representation. Given the mapping from 2D square to a spherical mesh and another from the sphere to our template shape, our mapping from 2D manifold to the template shape is their composition.  Constructing a mapping between 2D square and sphere can be understood to one analogous to latitudes and longitudes on the globe.

All our template shapes are genus-0 (isomorphic to a sphere -- without holes). They have been pre-processed to have 642 vertices and 1280 faces using Meshlab's ~\cite{LocalChapterEvents:ItalChap:ItalianChapConf2008:129-136} quadratic vertex decimation \cite{garland1997surface}. Constructing a  mapping between a sphere and a template shape corresponds to finding a mapping between faces of the spherical mesh and the faces of the template shape. To find such a mapping we need to deform the sphere to ensure that the corresponding faces have similar areas. We to do this by minimizing the squared difference of logarithm of triangle areas as the objective using Adam ~\cite{kingma2014adam} optimizer. This optimization is an offline process and is part of preprocessing for the given category.

\vspace{1mm}
\noindent
\textbf{Parametrizing Articulation.} Every template shape has a defined part hierarchy that assigns every node a parent except for the \emph{root} node. A rigid transform is represented as translation $t$ and rotation $R$. Every node, $k$ has a rigid transform associated with. This rigid transform is applied in the frame of the part. Consider $\mathcal{T}_{k}^{\prime}$ to be the local rigid transform at the node $k$ represented using $R$ and $t$. We define the global rigid transform at node $k$ as $\mathcal{T}_{k}$.
\begin{align}
    \mathcal{T}_{k} &= \mathcal{T}_{l} \circ \mathcal{T}_{k}^{\prime} ; \\\quad \quad &\text{where node $l$ is the parent of node $k$} \nonumber
\end{align}

\vspace{1mm}
\noindent
\textbf{Barycentric Interpolation.} We use barycentric interpolation to compute the point on the surface of of the mesh for every $u \in [0,1)^{2}$.  Given a $u$ we map the point to the surface of the mesh of sphere, and then find the face it belongs to. We use the vertices to this face to compute it's barycentric coordinates. We then use these computed barycentric coordinates and the vertices on the corresponding face of the articulated mesh to compute the 3D location of the point.

\section{Implementation Details}
Our implementation uses PyTorch ~\cite{paszke2017automatic}
\subsection{Network Details}
We use a Resnet18 ~\cite{he2016deep} encoder extracted with features after 4 blocks and 5 layer decoder. Our encoder is initialized with pretrained ImageNet \cite{deng2009imagenet} features. The encoder-decoder takes input an image and then outputs a 3D unit vector per pixel. We convert this unit-vector to a 2D coordinate $u \in [0,1)^2$ which is used to parameterize our 2D square.
\subsection{Optimization}
\noindent
{\bf Parameterizing Part Transforms.}
We parameterize part transforms as an axis, angle representation. Every part's axis serves as a bias in the network that is learned and is same across the whole category. We predict the angle using a deep network for every part $k$.\\
\noindent
{\bf Parameterizing Camera Pose.} We parameterize the camera as orthographic similar to \cite{cmrKanazawa18} where we predict $R$ as a unit quaternion $q \in \mathcal{R}^{4}$ , $s \in (0, \infty), t \in \mathcal{R}^{2}$. Also, similar to ~\cite{kulkarni2019canonical} we predictor 8 hypothesis for camera pose and part transforms to ease the leaning process.\\
\noindent
{\bf Training for Articulation.} We first train our network to only learn camera pose predictions (along with the CSM predictor) for 10000 iterations. We then allow the model to articulate for the rest of the training iteration. We train with Adam ~\cite{kingma2014adam} as the optimizer using a learning rate of $10^{-4}$.

\subsection{Losses}
\noindent
{\bf Mask Consistency.}
We want to ensure that the rendered mask $M_{\text{rendered}}$ of the mesh under camera $\pi$ lies inside the ground truth mask of the object. We compute it by computing the euclidean distance field $D_{\text{edf}}$ for the ground truth mask summed over all the pixels in the rendered mask.
\begin{align}
    L_{\text{mask-consistency}} = \sum_{p} M_{\text{rendered}}[p]  D_{\text{edf}}[p]
\end{align}
{\bf Mask Coverage.} Enforcing consistency is not enough as it only forces the object to lie inside the mask. We also want to ensure that all the  object pixels in the foreground should be close to re-projection of some mesh vertex. This loss competes against with the consistency loss objective.
\begin{align}
    L_{\text{mask-coverage}} =  \sum_{p  \in I_{f}} \min_{v \in V} \norm{v - p}
\end{align}
{\bf Mask Loss.}
Our mask objective is a sum of these two competing losses as follows.
\begin{align}
    L_{\text{mask}} = L_{\text{mask-consistency}}  + L_{\text{mask-coverage}}
\end{align}
{\bf Visibility Loss.}
We render depth map of the articulated template shape under camera $\pi$ as $D_{\pi}$ and for every pixel we have the $z_{p}$ as z-coordinate of the pixel in the camera frame corresponding to the point $\delta(\phi(C[p]))$ on the surface of the 3D shape.
\begin{align}
    L_{\text{vis}} = \sum_{p\in I_{f}} \max(0, z_{p} - D_{\pi}[\mathbf{\bar{p}}]); \quad \mathbf{\bar{p}}= \pi(\delta(\phi(C[p])))
\end{align}
{\bf Regularization Losses.} Additionally, we also add the regularization to translation prediction for part transforms, along with an entropy penalty to encourage diversity of the multi-pose predictor.

\section{Evaluation}
\subsection{Transfer Keypoints using CSM}
We use the predicted CSM map to transfer keypoints between source and target images by using their corresponding canonical surface mappings $C_{s}$ and $C_{t}$. For every source keypoint at pixel $p^{k}$ we map the keypoint to the point the non-articulated template shape, and then search for a pixel on the target image that maps closest to this point. 
\begin{align}
    T_{s\xrightarrow{}t}^{k} = \argmin_{p} (\norm{\phi(C_{t}[p]) - \phi(C_{s}[p^{k}])})
\end{align}

\begin{table}[t!]
\caption{ We evaluate the performance of our model trained with  ground truth (GT) and Mask-RCNN segmentation on the task of Keypoint (KP) Transfer and Keypoint (KP) Projection as described in Table 1 and 2 of the main text. We report the performance on the CUBS-2011 \cite{wah2011caltech} dataset, and observe that there is not a significant performance gap when using segmentations from Mask-RCNN.
}
\tablelabel{pcktransfermrcnn}
\setlength{\tabcolsep}{4pt}

\centering 
% \small{
% \scalebox{1.0}{
\resizebox{
        \width
      }{!}{
\begin{tabular}{l|rr} \toprule
% \multicolumn{2}{c}{Item} \\ \cmidrule(r){1-2}
Mask Source & PCK Transfer & KP Projection\\ \midrule
GT (Humans)       & 42.6 & 46.8  \\
Mask-RCNN~\cite{he2017mask}     & 38.5 & 45.5 \\
 \bottomrule
\end{tabular}
}

\end{table}

\subsection{Importance of Ground Truth Masks}
We study the impact of having foreground segmentation from Mask-RCNN versus using human annotation for \emph{training} A-CSM model. In these experiments we use pre-trained Mask-RCNN \cite{he2017mask} on 80 COCO \cite{lin2014microsoft} categories. We use the CUB-2011 \cite{wah2011caltech} dataset with segmentation from Mask-RCNN ~\cite{he2017mask} and the ground truth annotations to compared the performance on tasks of PCK-Transfer and Key-Point (KP) Projection. We report results in \tableref{pcktransfermrcnn} and observe that though our method has a superior performance with precise ground truth masks, the performance drop with using automatically generated inaccurate segmentation from Mask-RCNN is not significant and our methods remains competitive.

\vspace{1mm}
\noindent
\textbf{Evaluation using GT Masks.} Our results in the main paper for PCK Transfer use predicted masks to transfer keypoints between two given images. We evaluated the performance of transfer by using ground truth masks on the birds dataset and observe that the performance changes on an average of 0.1 points. This implies that there is no significant disadvantage in using the predicted masks for evaluation. 
\section{Qualitative Sampled Results on 11 Categories}
\begin{table*}[!t]
\setlength{\tabcolsep}{0.02em}
\renewcommand{\arraystretch}{1}
\centering
  \scalebox{0.75}{
\begin{tabular}{lrl@{\hskip 0.05em}rl@{\hskip 0.05em}rl@{\hskip 0.05em}rl@{\hskip 0.05em}rl}
\addpic{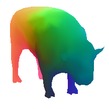} &
\addpic{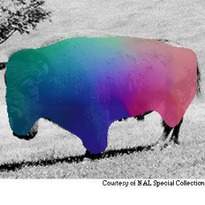} &
\addpichalf{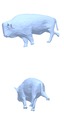} &
\addpic{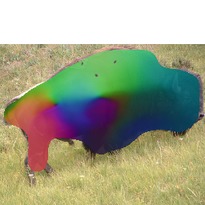} &
\addpichalf{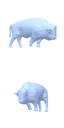} &
\addpic{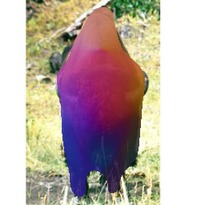} &
\addpichalf{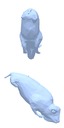} &
\addpic{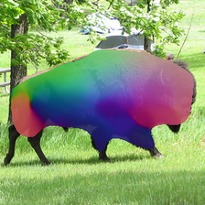} &
\addpichalf{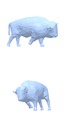} \\  
 & \addpic{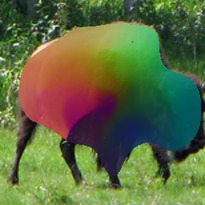} &
\addpichalf{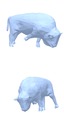} &
\addpic{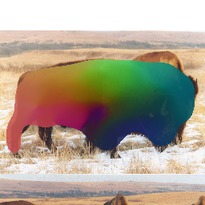} &
\addpichalf{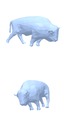} &
\addpic{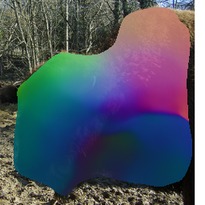} &
\addpichalf{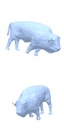} &
\addpic{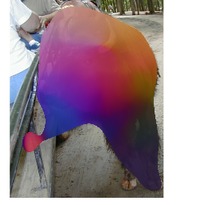} &
\addpichalf{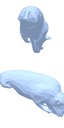} \\  
 & \addpic{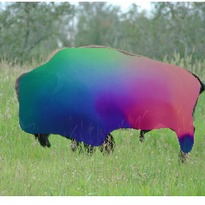} &
\addpichalf{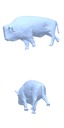} &
\addpic{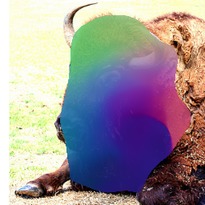} &
\addpichalf{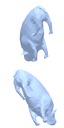} &
\addpic{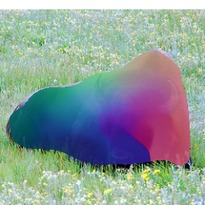} &
\addpichalf{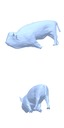} &
\addpic{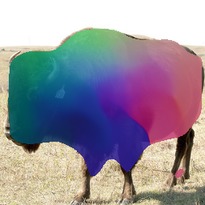} &
\addpichalf{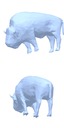} \\  
% & \addpic{figures/supp/bison/21/ind_21_uv_overlay.jpg} &
% \addpichalf{figures/supp/bison/21/ind_21_pose.jpg} &
% \addpic{figures/supp/bison/84/ind_84_uv_overlay.jpg} &
% \addpichalf{figures/supp/bison/84/ind_84_pose.jpg} &
% \addpic{figures/supp/bison/99/ind_99_uv_overlay.jpg} &
% \addpichalf{figures/supp/bison/99/ind_99_pose.jpg} &
% \addpic{figures/supp/bison/88/ind_88_uv_overlay.jpg} &
% \addpichalf{figures/supp/bison/88/ind_88_pose.jpg}  \\ 
%%%%%%%%%%%%%%%%%%%%%%%%%%%%%%%%%%%%
\midrule
\addpic{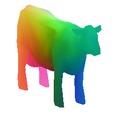} &
\addpic{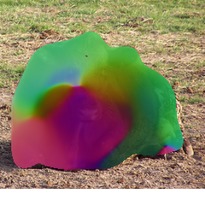} &
\addpichalf{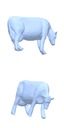} &
\addpic{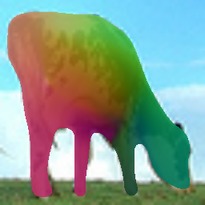} &
\addpichalf{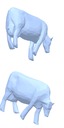} &
\addpic{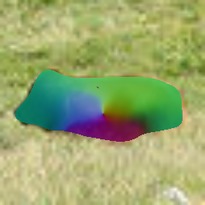} &
\addpichalf{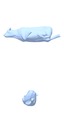} &
\addpic{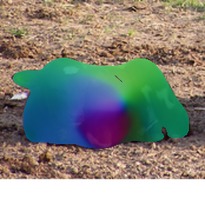} &
\addpichalf{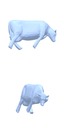} \\  
 & \addpic{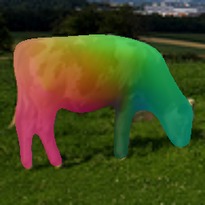} &
\addpichalf{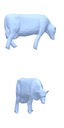} &
\addpic{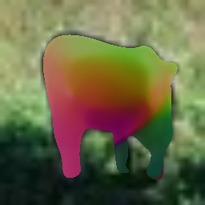} &
\addpichalf{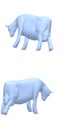} &
\addpic{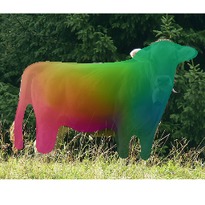} &
\addpichalf{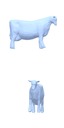} &
\addpic{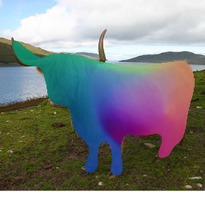} &
\addpichalf{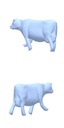} \\  
 & \addpic{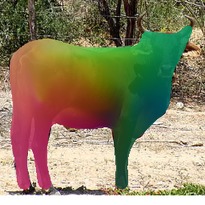} &
\addpichalf{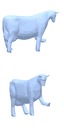} &
\addpic{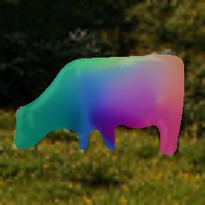} &
\addpichalf{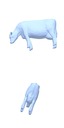} &
\addpic{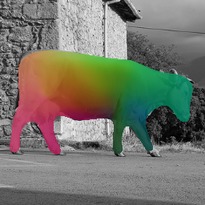} &
\addpichalf{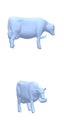} &
\addpic{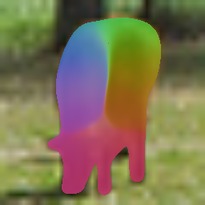} &
\addpichalf{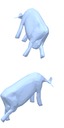} \\  
% & \addpic{figures/supp/cow_8parts/100/ind_100_uv_overlay.jpg} &
% \addpichalf{figures/supp/cow_8parts/100/ind_100_pose.jpg} &
% \addpic{figures/supp/cow_8parts/80/ind_80_uv_overlay.jpg} &
% \addpichalf{figures/supp/cow_8parts/80/ind_80_pose.jpg} &
% \addpic{figures/supp/cow_8parts/58/ind_58_uv_overlay.jpg} &
% \addpichalf{figures/supp/cow_8parts/58/ind_58_pose.jpg} &
% \addpic{figures/supp/cow_8parts/11/ind_11_uv_overlay.jpg} &
% \addpichalf{figures/supp/cow_8parts/11/ind_11_pose.jpg} \\
\midrule
\addpic{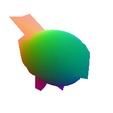} &
\addpic{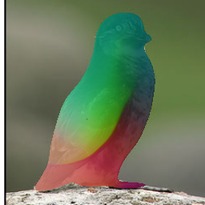} &
\addpichalf{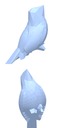} &
\addpic{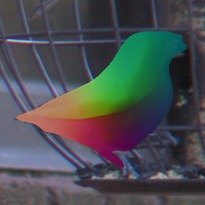} &
\addpichalf{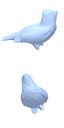} &
\addpic{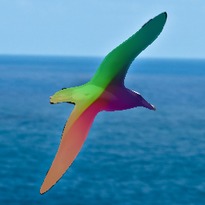} &
\addpichalf{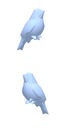} &
\addpic{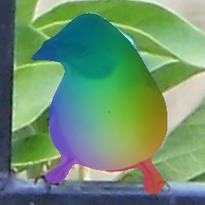} &
\addpichalf{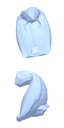} \\  
 & \addpic{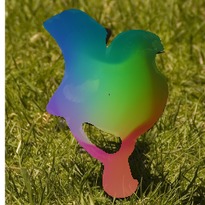} &
\addpichalf{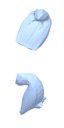} &
\addpic{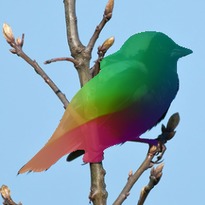} &
\addpichalf{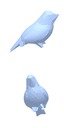} &
\addpic{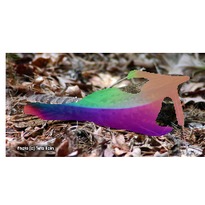} &
\addpichalf{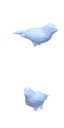} &
\addpic{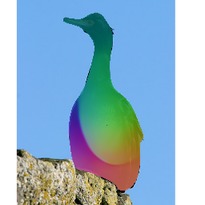} &
\addpichalf{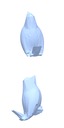} \\  
 & \addpic{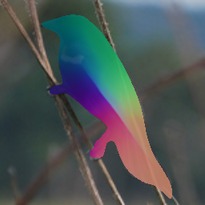} &
\addpichalf{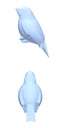} &
\addpic{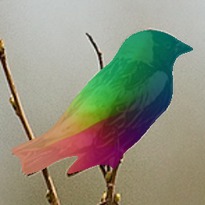} &
\addpichalf{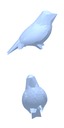} &
\addpic{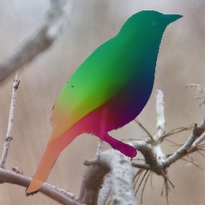} &
\addpichalf{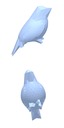} &
\addpic{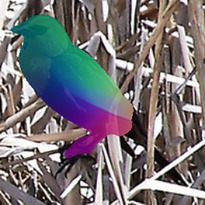} &
\addpichalf{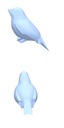} \\
\bottomrule \\
\end{tabular}
}
\vspace{-4mm}
\captionof{figure}{
{\emph{Randomly sampled} results on bisons, cows, and birds}
}
\figlabel{qual1}
\end{table*} 
\begin{table*}[!t]
\setlength{\tabcolsep}{0.02em}
\renewcommand{\arraystretch}{1}
\centering
  \scalebox{0.75}{
\begin{tabular}{lrl@{\hskip 0.05em}rl@{\hskip 0.05em}rl@{\hskip 0.05em}rl@{\hskip 0.05em}rl}
\addpic{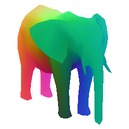} &
\addpic{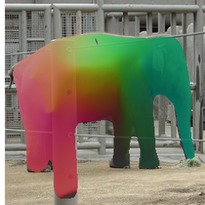} &
\addpichalf{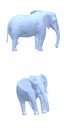} &
\addpic{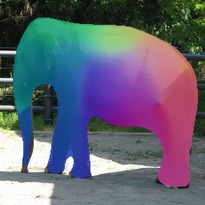} &
\addpichalf{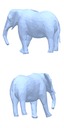} &
\addpic{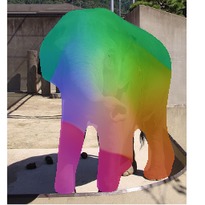} &
\addpichalf{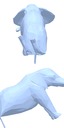} &
\addpic{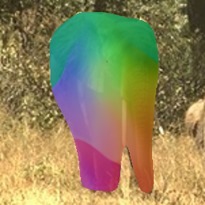} &
\addpichalf{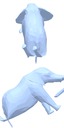} \\  
 & \addpic{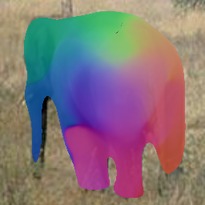} &
\addpichalf{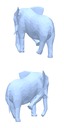} &
\addpic{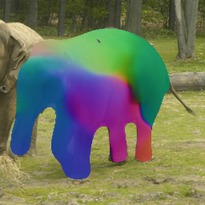} &
\addpichalf{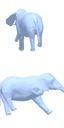} &
\addpic{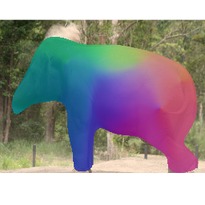} &
\addpichalf{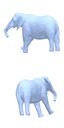} &
\addpic{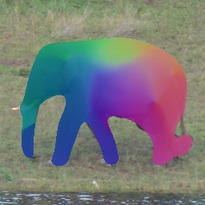} &
\addpichalf{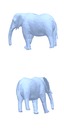} \\  
 & \addpic{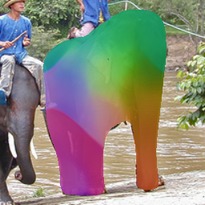} &
\addpichalf{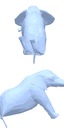} &
\addpic{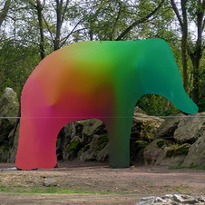} &
\addpichalf{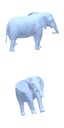} &
\addpic{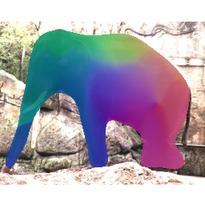} &
\addpichalf{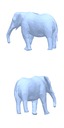} &
\addpic{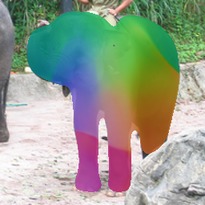} &
\addpichalf{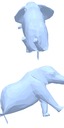} \\  
\midrule
\addpic{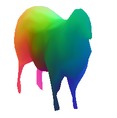} &
\addpic{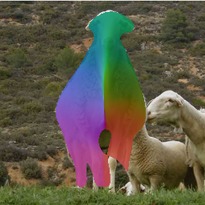} &
\addpichalf{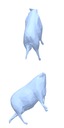} &
\addpic{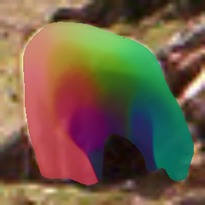} &
\addpichalf{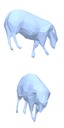} &
\addpic{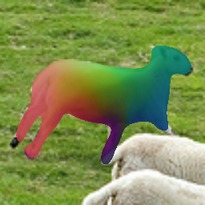} &
\addpichalf{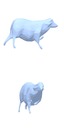} &
\addpic{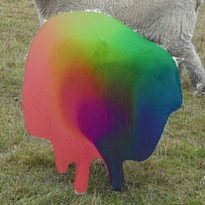} &
\addpichalf{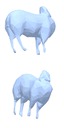} \\  
 & \addpic{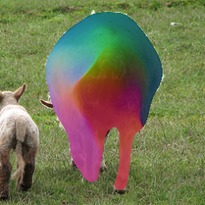} &
\addpichalf{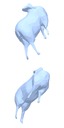} &
\addpic{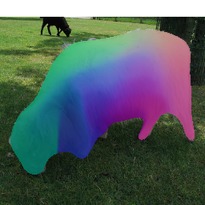} &
\addpichalf{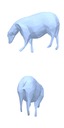} &
\addpic{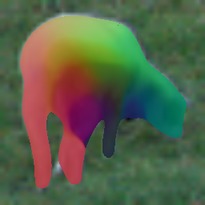} &
\addpichalf{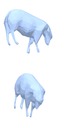} &
\addpic{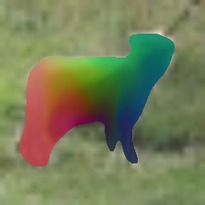} &
\addpichalf{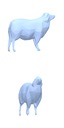} \\  
 & \addpic{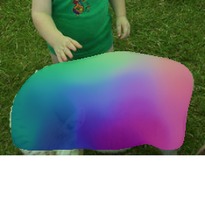} &
\addpichalf{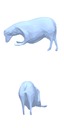} &
\addpic{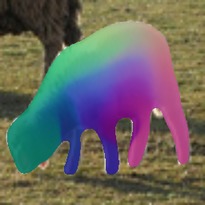} &
\addpichalf{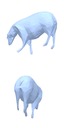} &
\addpic{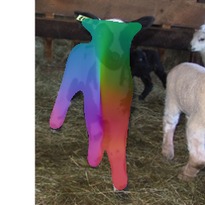} &
\addpichalf{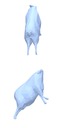} &
\addpic{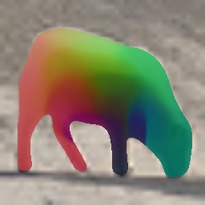} &
\addpichalf{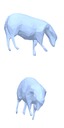} \\
\bottomrule \\
\addpic{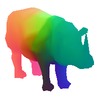} &
\addpic{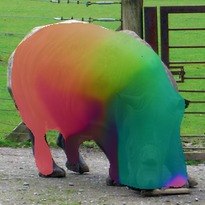} &
\addpichalf{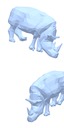} &
\addpic{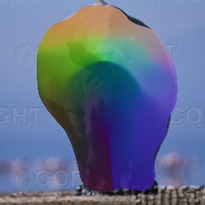} &
\addpichalf{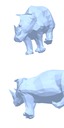} &
\addpic{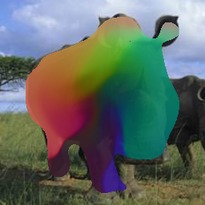} &
\addpichalf{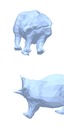} &
\addpic{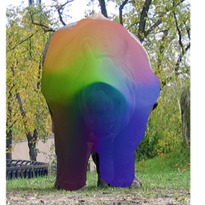} &
\addpichalf{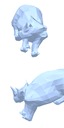} \\  
 & \addpic{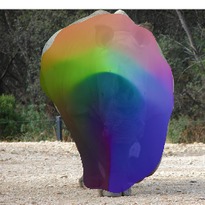} &
\addpichalf{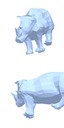} &
\addpic{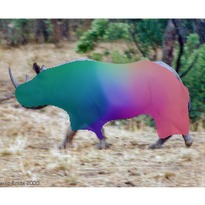} &
\addpichalf{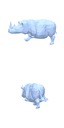} &
\addpic{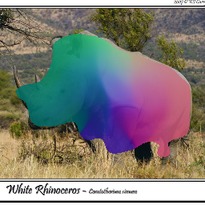} &
\addpichalf{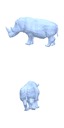} &
\addpic{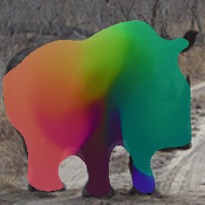} &
\addpichalf{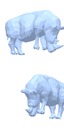} \\  
 & \addpic{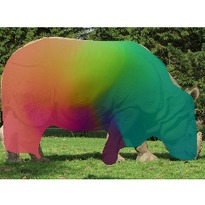} &
\addpichalf{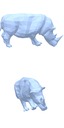} &
\addpic{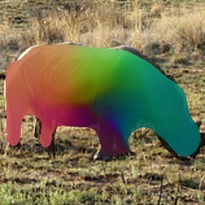} &
\addpichalf{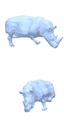} &
\addpic{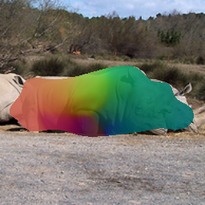} &
\addpichalf{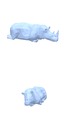} &
\addpic{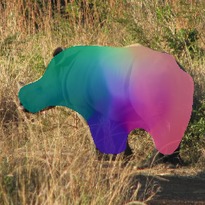} &
\addpichalf{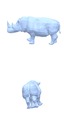} \\ 
\end{tabular}
}
\vspace{-4mm}
\captionof{figure}{
{\emph{Randomly sampled} results on elephants, sheeps, and rhinos}
%The figure show articulations of template shape for every input image along side the CSM prediction for the foreground pixels. We observe consistent CSM predictions for various functional regions of the object. For instance, we can see the head of all the quadrupeds is greenish in color which accurately represents its mapping to green region on the template shape shown in the right-most column. We show results over 11 categories with a wide variety of articulations, but in certain cases we observe the model compensating by articulating excessively while sometimes under articulating.
}
\figlabel{qual2}
\end{table*} 
\begin{table*}[!t]
\setlength{\tabcolsep}{0.02em}
\renewcommand{\arraystretch}{1}
\centering
  \scalebox{0.75}{
\begin{tabular}{lrl@{\hskip 0.05em}rl@{\hskip 0.05em}rl@{\hskip 0.05em}rl@{\hskip 0.05em}rl}
\addpic{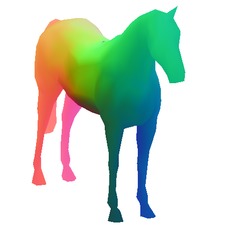} &
\addpic{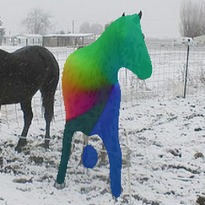} &
\addpichalf{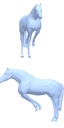} &
\addpic{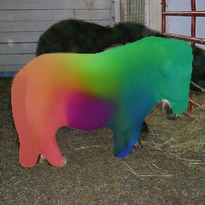} &
\addpichalf{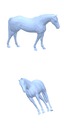} &
\addpic{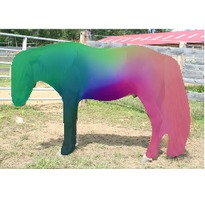} &
\addpichalf{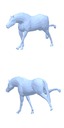} &
\addpic{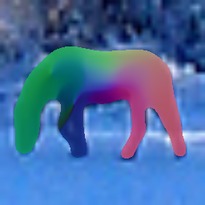} &
\addpichalf{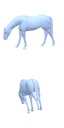} \\  
 & \addpic{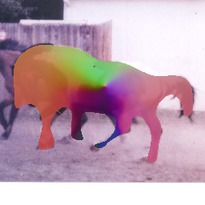} &
\addpichalf{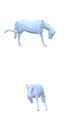} &
\addpic{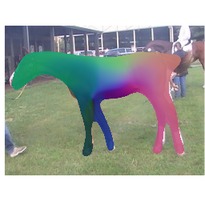} &
\addpichalf{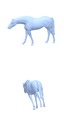} &
\addpic{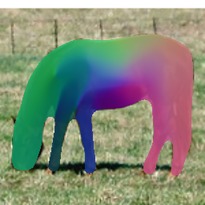} &
\addpichalf{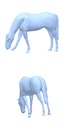} &
\addpic{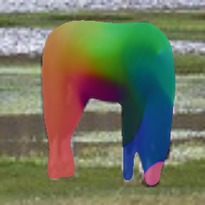} &
\addpichalf{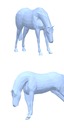} \\  
 & \addpic{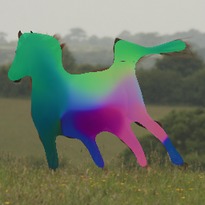} &
\addpichalf{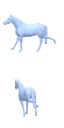} &
\addpic{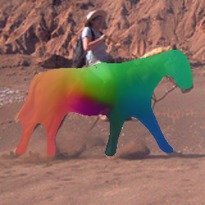} &
\addpichalf{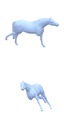} &
\addpic{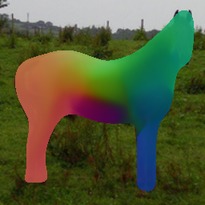} &
\addpichalf{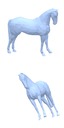} &
\addpic{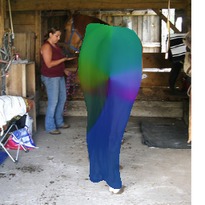} &
\addpichalf{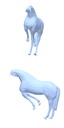} \\
\midrule
\addpic{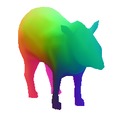} &
\addpic{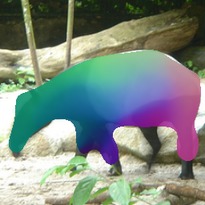} &
\addpichalf{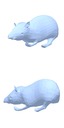} &
\addpic{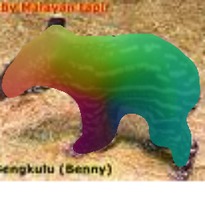} &
\addpichalf{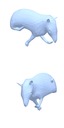} &
\addpic{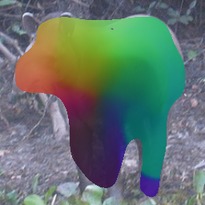} &
\addpichalf{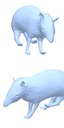} &
\addpic{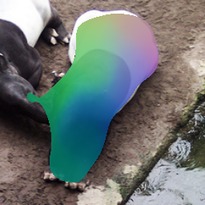} &
\addpichalf{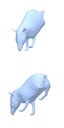} \\  
 & \addpic{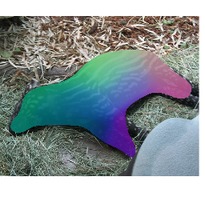} &
\addpichalf{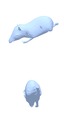} &
\addpic{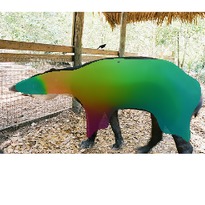} &
\addpichalf{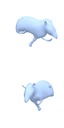} &
\addpic{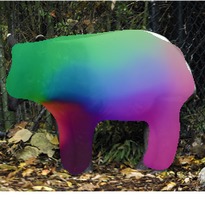} &
\addpichalf{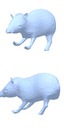} &
\addpic{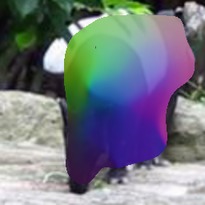} &
\addpichalf{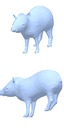} \\  
 & \addpic{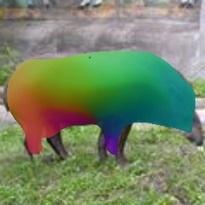} &
\addpichalf{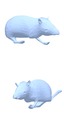} &
\addpic{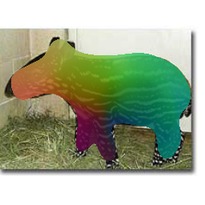} &
\addpichalf{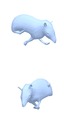} &
\addpic{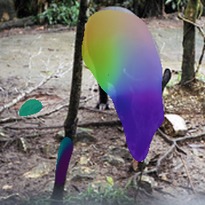} &
\addpichalf{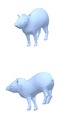} &
\addpic{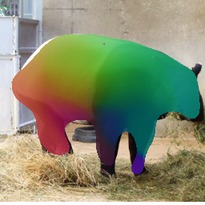} &
\addpichalf{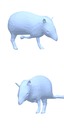} \\
%%%%%%%%%%%%%%%%%%%%%%%%%%%%%%%%%%%%
\midrule
\addpic{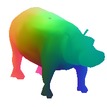} &
\addpic{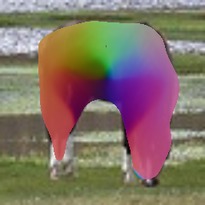} &
\addpichalf{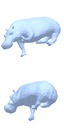} &
\addpic{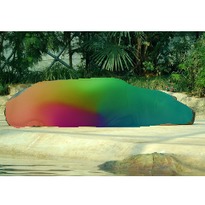} &
\addpichalf{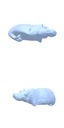} &
\addpic{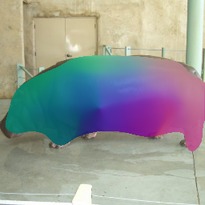} &
\addpichalf{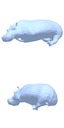} &
\addpic{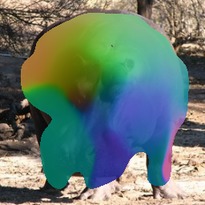} &
\addpichalf{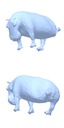} \\  
 & 
\addpic{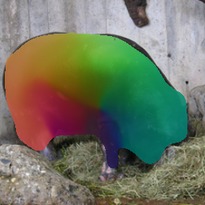} &
\addpichalf{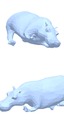} &
\addpic{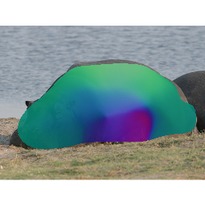} &
\addpichalf{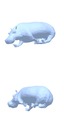}&
\addpic{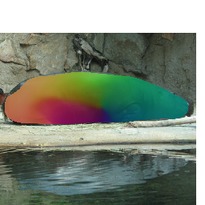} &
\addpichalf{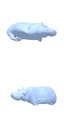} &
\addpic{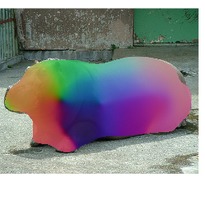} &
\addpichalf{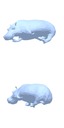} \\  
 & \addpic{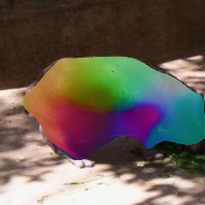} &
\addpichalf{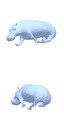} &
\addpic{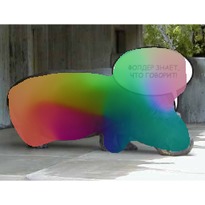} &
\addpichalf{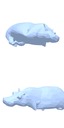}   &
\addpic{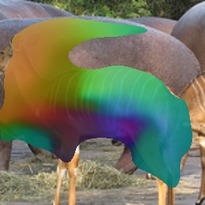} &
\addpichalf{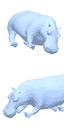} &
\addpic{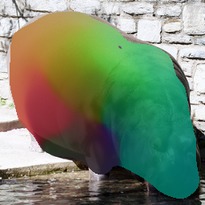} &
\addpichalf{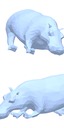} \\  
\bottomrule \\
\end{tabular}
}
\vspace{-4mm}
\captionof{figure}{
{\emph{Randomly sampled} results on horses, tapirs, and hippos}
%The figure show articulations of template shape for every input image along side the CSM prediction for the foreground pixels. We observe consistent CSM predictions for various functional regions of the object. For instance, we can see the head of all the quadrupeds is greenish in color which accurately represents its mapping to green region on the template shape shown in the right-most column. We show results over 11 categories with a wide variety of articulations, but in certain cases we observe the model compensating by articulating excessively while sometimes under articulating.
}
\figlabel{qual3}
\end{table*} 
\begin{table*}[!t]
\setlength{\tabcolsep}{0.02em}
\renewcommand{\arraystretch}{1}
\centering
  \scalebox{0.75}{
\begin{tabular}{lrl@{\hskip 0.05em}rl@{\hskip 0.05em}rl@{\hskip 0.05em}rl@{\hskip 0.05em}rl}
\addpic{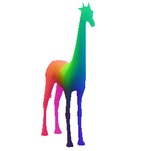} &
\addpic{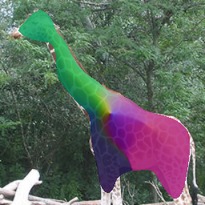} &
\addpichalf{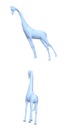} &
\addpic{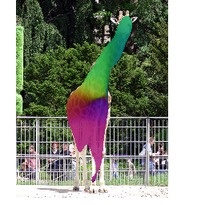} &
\addpichalf{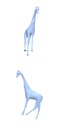} &
\addpic{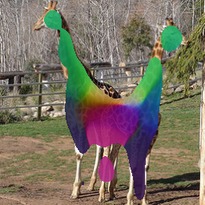} &
\addpichalf{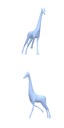} &
\addpic{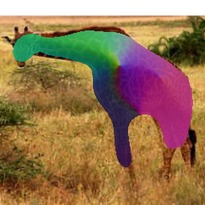} &
\addpichalf{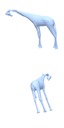} \\  
 & \addpic{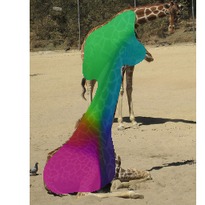} &
\addpichalf{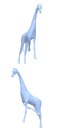} &
\addpic{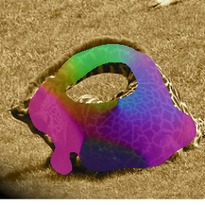} &
\addpichalf{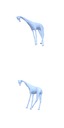} &
\addpic{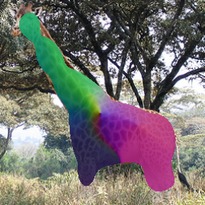} &
\addpichalf{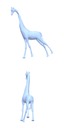} &
\addpic{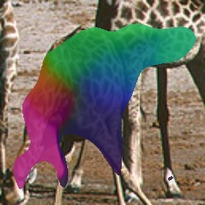} &
\addpichalf{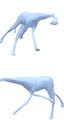} \\  
 & \addpic{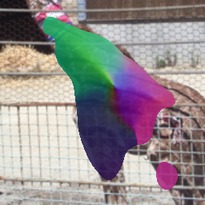} &
\addpichalf{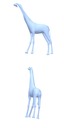} &
\addpic{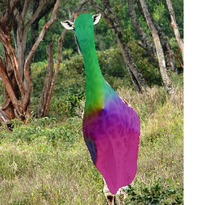} &
\addpichalf{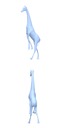} &
\addpic{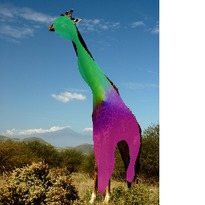} &
\addpichalf{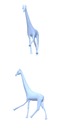} &
\addpic{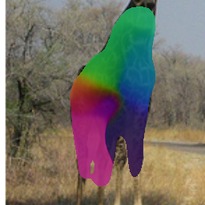} &
\addpichalf{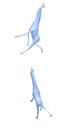} \\ 
\midrule
\addpic{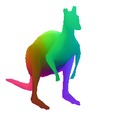} &
\addpic{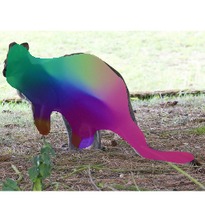} &
\addpichalf{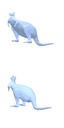} &
\addpic{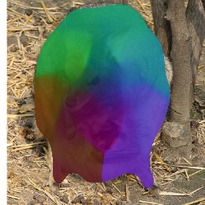} &
\addpichalf{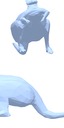} &
\addpic{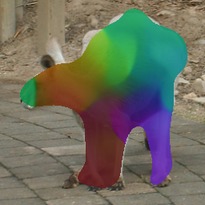} &
\addpichalf{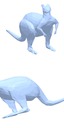} &
\addpic{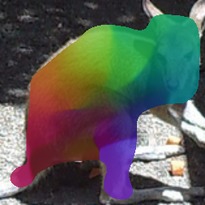} &
\addpichalf{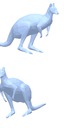} \\  
 & \addpic{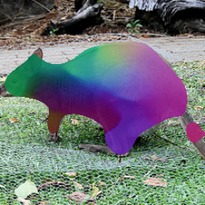} &
\addpichalf{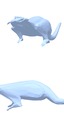} &
\addpic{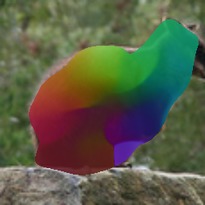} &
\addpichalf{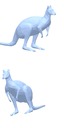} &
\addpic{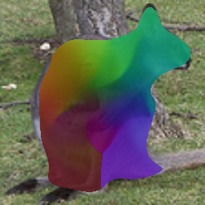} &
\addpichalf{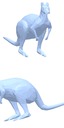} &
\addpic{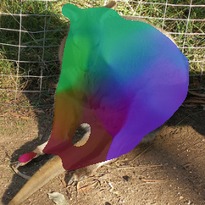} &
\addpichalf{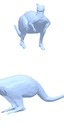} \\  
 & \addpic{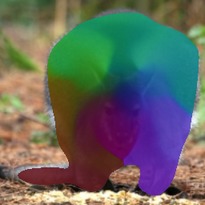} &
\addpichalf{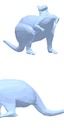} &
\addpic{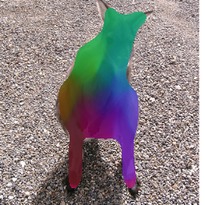} &
\addpichalf{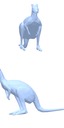} &
\addpic{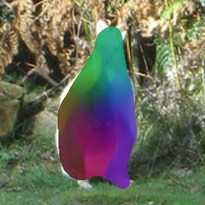} &
\addpichalf{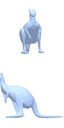} &
\addpic{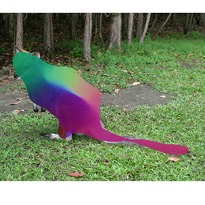} &
\addpichalf{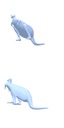} \\
%%%%%%%%%%%%%%%%%%%%%%%%%%%%%%%%%%%%
\bottomrule \\
\end{tabular}
}
\vspace{-4mm}
\captionof{figure}{
{\emph{Randomly sampled} results on giraffes and kangaroos}
%The figure show articulations of template shape for every input image along side the CSM prediction for the foreground pixels. We observe consistent CSM predictions for various functional regions of the object. For instance, we can see the head of all the quadrupeds is greenish in color which accurately represents its mapping to green region on the template shape shown in the right-most column. We show results over 11 categories with a wide variety of articulations, but in certain cases we observe the model compensating by articulating excessively while sometimes under articulating.
}
\figlabel{qual4}
\end{table*} 
We randomly sample results for all the categories shown in the paper show their visualizations in
\figref{qual1}, \figref{qual2}, \figref{qual3}, \figref{qual4}. These figure show articulations of template shape for every input image along side the CSM prediction for the foreground pixels. We observe consistent CSM predictions for various functional regions of the object. For instance, we can see the head of all the quadrupeds is greenish in color which accurately represents its mapping to green region on the template shape shown in the right-most column. We show results over 11 categories with a wide variety of articulations. %, but in certain cases we observe the model compensating by articulating excessively while sometimes under articulating.

We also show results of our method on a few videos downloaded from the internet in the video file submitted along with this supplementary. Our method is applied on a per frame basis without any temporal smoothing. We show a few screenshots from the video in  \figref{video}

\begin{figure*}
    \centering
    \includegraphics[width=0.40\textwidth]{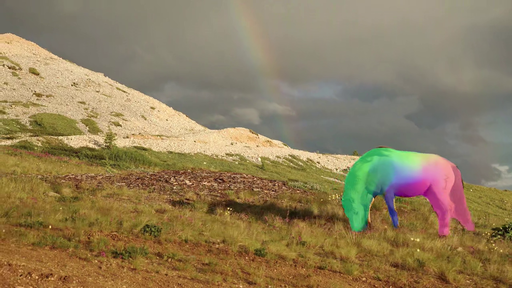}
    \includegraphics[width=0.40\textwidth]{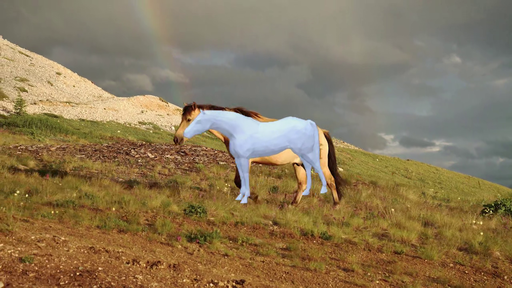}
    \includegraphics[width=0.40\textwidth]{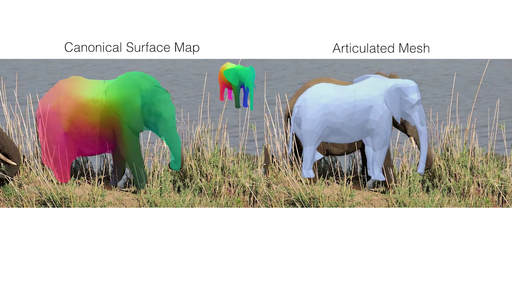}
    \includegraphics[width=0.40\textwidth]{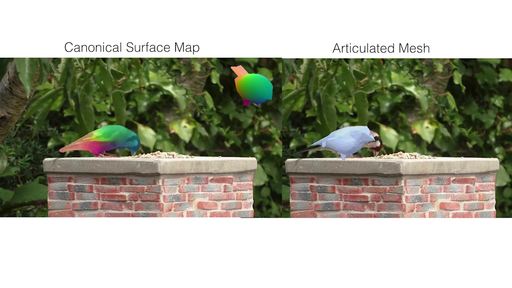}
    \includegraphics[width=0.40\textwidth]{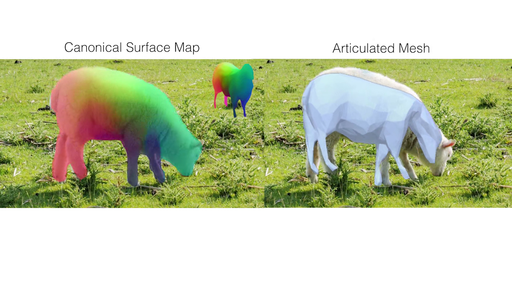}
    \includegraphics[width=0.40\textwidth]{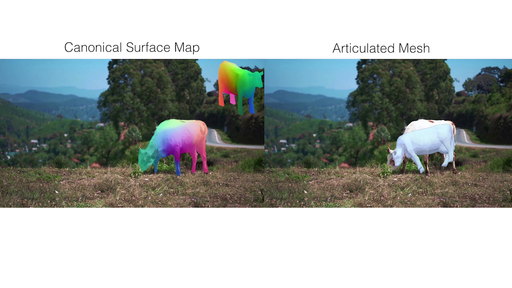}
    \caption{Screenshots of results from our method on the videos from the Internet}
    \figlabel{video}
\end{figure*}

\end{document}